\documentclass[pmlr]{jmlr}

\RequirePackage{graphicx}
\usepackage{booktabs}
\usepackage{longtable}
\makeatletter
\def\set@curr@file#1{\def\@curr@file{#1}} 
\makeatother
\usepackage[load-configurations=version-1]{siunitx} 

\usepackage{enumitem}
\usepackage{caption}
\usepackage{multirow}

\newcommand{\vx}{\boldsymbol{x}}
\newcommand{\vz}{\boldsymbol{z}}
\newcommand{\vw}{\boldsymbol{w}}
\newcommand{\vs}{\boldsymbol{s}}
\newcommand{\cL}{\mathcal{L}}
\theorembodyfont{\upshape}
\theoremheaderfont{\scshape}
\theorempostheader{:}
\theoremsep{\newline}

\jmlrvolume{298}
\jmlryear{2025}
\jmlrworkshop{Machine Learning for Healthcare}

\title[Borrowing From the Future]{Borrowing From the Future: Enhancing Early Risk Assessment through Contrastive Learning}

\newcommand{\addrdukebnb}{Department of Biostatistics \& Bioinformatics, Duke University,
Durham, NC, USA}

\author{\Name{Minghui Sun}$^{1}$ \Email{minghui.sun@duke.edu} 
\AND \Name{Matthew M. Engelhard}$^{1}$ \Email{m.engelhard@duke.edu} 
\AND \Name{Benjamin A. Goldstein}$^{1}$ \Email{ben.goldstein@duke.edu}
\AND $^{1}$ \addr \addrdukebnb }

\begin{document}
    \maketitle

    \begin{abstract}
        Risk assessments for a pediatric population are often conducted across multiple stages. For example, clinicians may evaluate risks prenatally, at birth, and during Well-Child visits. Although predictions made at later stages typically achieve higher precision, it is clinically desirable to make reliable risk assessments as early as possible. Therefore, this study focuses on improving prediction performance in early-stage risk assessments. Our solution, \textbf{Borrowing From the Future (BFF)}, is a contrastive multi-modal framework that treats each time window as a distinct modality. In BFF, a model is trained on all available data throughout the time while performing a risk assessment using up-to-date information. This contrastive framework allows the model to ``borrow'' informative signals from later stages (e.g., Well-Child visits) to implicitly supervise the learning at earlier stages (e.g., prenatal/birth stages). We validate BFF on two real-world pediatric outcome prediction tasks, demonstrating consistent improvements in early risk assessments. The code is available
        at \url{https://github.com/scotsun/bff}.
    \end{abstract}

    \section{Introduction}

    Clinical risk assessments can be conducted at multiple points in time. For a pediatric population, risk may be assessed prenatally, at birth, and during developmental Well-Child visits \citep{lipkin2020promoting}. Numerous studies have leveraged prenatal information to develop clinical prediction models for pediatric outcomes, highlighting important links between maternal prenatal factors and child health \citep{beijers2010maternal, walsh2019maternal, kong2024predicting}.  Risk assessments performed at later time points tend to be more accurate because: (i) additional information has accumulated over watchful waiting, and (ii) the child is temporally closer to the potential onset of the condition, making recent data more predictive of the eventual diagnosis. However, it is clinically desirable to perform risk assessments as early as possible, allowing timely interventions and preventive actions. In this work, our goal is to improve the risk assessment at earlier points in time.

    A clinical prediction model (CPM) that relies solely on the data from the earlier time windows may have limited performance, since the data may lack explicit and direct predictive relevance to the target outcome. Therefore, we develop a contrastive learning framework to adjust the learning of the early information. In general, contrastive learning improves representation quality by leveraging information from different views of the same sample \citep{chen2020simple, tian2020contrastive}, or by integrating data across different modalities observed from the same sample \citep{radford2021learning}. In existing contrastive methods for sequential data, representations are learned either at the overall sequence level or at individual timestamps \citep{gao2021simcse, yue2022ts2vec, lee2023soft}. However, for our clinical use case, where risk assessments are performed across distinct time windows, we aim to efficiently learn representations at the time window level.

    To achieve our objective, we introduce \textbf{Borrowing From the Future (BFF)}, a contrastive multi-modal framework that reinterprets modality in a different way. While traditional multi-modal approaches integrate heterogeneous data types (e.g., text and images) to leverage complementary information, our framework takes a different approach. We treat each temporal segment (such as maternal prenatal, at-birth, and child developmental stages) as a distinct modality, thereby converting the single-modality medical code data into a multi-modal configuration.

    One key element in the BFF framework is the contrastive regularization (CR). Drawing on advances from unsupervised and self-supervised contrastive learning studies \citep{oord2018representation, tian2020contrastive, radford2021learning, yuan2021multimodal, lin2022multimodal}, our CR contains both within-modal and across-modal contrastive losses. This dual strategy supports the learning of distinct temporal features. More importantly, training with across-modal alignment allows early modalities to benefit from stronger predictive signals found in later time points, effectively ``borrowing'' future information as implicit supervision to enhance patient-specific representations. The principle of supervising less informative modalities using a stronger modality is inspired by the design of CLIP \citep{radford2021learning}, which aligns visual and textual representations through cross-modal supervision.

    An effective way to aggregate multi-modal representations is crucial in multi-modal learning. We adapt the self-gating mechanism in SE block proposed by \cite{hu2018SEnet} to our multi-modal setting, replacing the original sigmoid gate with softmax gates for different modalities. We term this approach \textbf{Softmax Self-Gating}. This mechanism applies self-modulated softmax scores to individually weight the features from each modality before the final aggregation. Modalities can be easily masked off in the softmax to handle missing data or prevent access to future information during model evaluation. Meanwhile, the gating mechanism provides inherent model explainability, because the softmax scores provide an interpretable measure of the feature-wise attention given to each modality embedding in the final prediction task.

    To summarize, in the BFF framework, we treat Electronic Health Records (EHR) medical codes observed from different time windows as separate modalities. We use contrastive learning to separately extract modality-specific and patient-specific features. During the learning of patient-specific features, the CR effectively borrows information from the future modality to improve the representation learning in the earlier modalities. We use Softmax Self-Gating to aggregate the multi-modal features. We use two real-world pediatric outcome prediction tasks to show the effectiveness of our framework: a time-to-diagnosis prediction task for Autism Spectrum Disorder (ASD) and a binary outcome prediction task of Recurrent Acute Otitis Media (rAOM). According to the experiment results of these two tasks, BFF improves a CPM's performance at early risk assessments.

    Our contributions are as follows:
    \begin{itemize}[itemsep=0pt, parsep=0pt, topsep=0pt]
        \item We propose BFF, a framework that converts uni-modal EHR data to multi-modal data by considering data over different time windows as separate modalities. It leverages contrastive learning to enhance CPMs' performance at early risk assessments.

        \item We design a contrastive regularization that jointly optimizes within-modal feature understanding and cross-modal representation alignment. The cross-modal alignment, in turn, enriches the representations learned on earlier modalities.

        \item We introduce Softmax Self-Gating, a straightforward yet powerful mechanism for achieving efficient and explainable multi-modal fusion.
    \end{itemize}

    \subsection*{Generalizable Insights about Machine Learning in the Context of
    Healthcare}
    Contrastive learning has been extensively explored as an effective approach to enhancing representation learning. In this study, we employ this technique to boost the performance of a CPM in early risk assessments. Data gathered during the child development period are highly relevant to clinical outcomes of interest. We leverage these data as a form of ``implicit'' supervision to guide the learning of historical information from the maternal prenatal period and at-birth encounter --- two clinically significant time windows for early risk assessments. Our proposed framework, BFF, is therefore capable of advancing preventive care for pediatric populations.

    \section{Related Work}

    \textbf{Multi-modal Contrastive Learning.} InfoNCE \citep{oord2018representation} is a foundational work for contrastive learning and is then extended to various settings. In self-supervised learning (SSL) settings, representations are learnt through \textit{within-modal alignment} \citep{chen2020simple, he2020momentum, gao2021simcse}. Different random augmentations applied to the same sample yield positive pairs, while augmentations of separate samples are used to construct negative pairs. In multi-modal learning settings, various streamlined methods learn multi-modal representations through cross-modal alignment \citep{tian2020contrastive, radford2021learning, qiu2023not, hager2023best}. In these scenarios, a positive pair refers to a matched pair of modalities from the same sample (e.g., a matched image-text pair) while a negative pair refers to a mismatched pair of modalities. Several recent works extend multi-modal contrastive learning with simultaneous within-modal and cross-modal objectives \citep{yuan2021multimodal, lin2022multimodal}. The integration strengthens modality-specific representation learning, and cross-modal interactions amplify the quality of the extracted features \citep{yuan2021multimodal}.

    \textbf{Contrastive Learning in Sequential Data.} TS2Vec \citep{yue2022ts2vec} and SoftCLT \citep{lee2023soft} learn contrastive representations for multi- and univariate time series in continuous space, and are typically applied to standard time series modalities. Contrastive learning is performed at two levels using augmented views of the sequences: an \textit{instance-wise} loss, which contrasts representations from different sequence instances at the same time point, and a \textit{temporal} loss, which contrasts representations from different timestamps within a single sequence. Both loss terms operate on the same representation vector. In contrast, BFF is applied to tokenized medical event sequences, which are structured similarly to NLP data. Contrastive learning is applied at two levels: time window and instance. The representation vector comprises two components, each governed by a distinct contrastive loss function. One targets \textit{within-modal} temporal dynamics, and the other facilitates \textit{across-modal} alignment at the instance level.

    \textbf{Attention and Soft Gating.} Soft gating techniques have been used in various works to improve the recurrent neural networks \citep{hochreiter1997long, cho2014learning, dauphin2017language}. These methods use input-dependent information control on features to address the vanishing gradient issue and improve representation learning. Similarly, SE block, leveraging sigmoid self-gating mechanism, recalibrates the feature map channel-wisely in convolutional neural networks \citep{hu2018SEnet}. These gating techniques improves feature discriminability, empowering the model to adaptively focus more on the important features but less on the trivial features. The self-attention mechanism introduced in Transformer also share the same functionality of adaptively control the feature expressiveness \citep{vaswani2017attention}. The attention weights operate similarly to soft gating weights, dynamically modulating the value embedding features through pairwise interactions between embeddings. This selective suppression and amplification of feature contributions generates contextually-aware representations.

    \section{Methods}

    \subsection{Overview of the BFF}

    \begin{figure}
        \centering
        \includegraphics[width=0.90\linewidth]{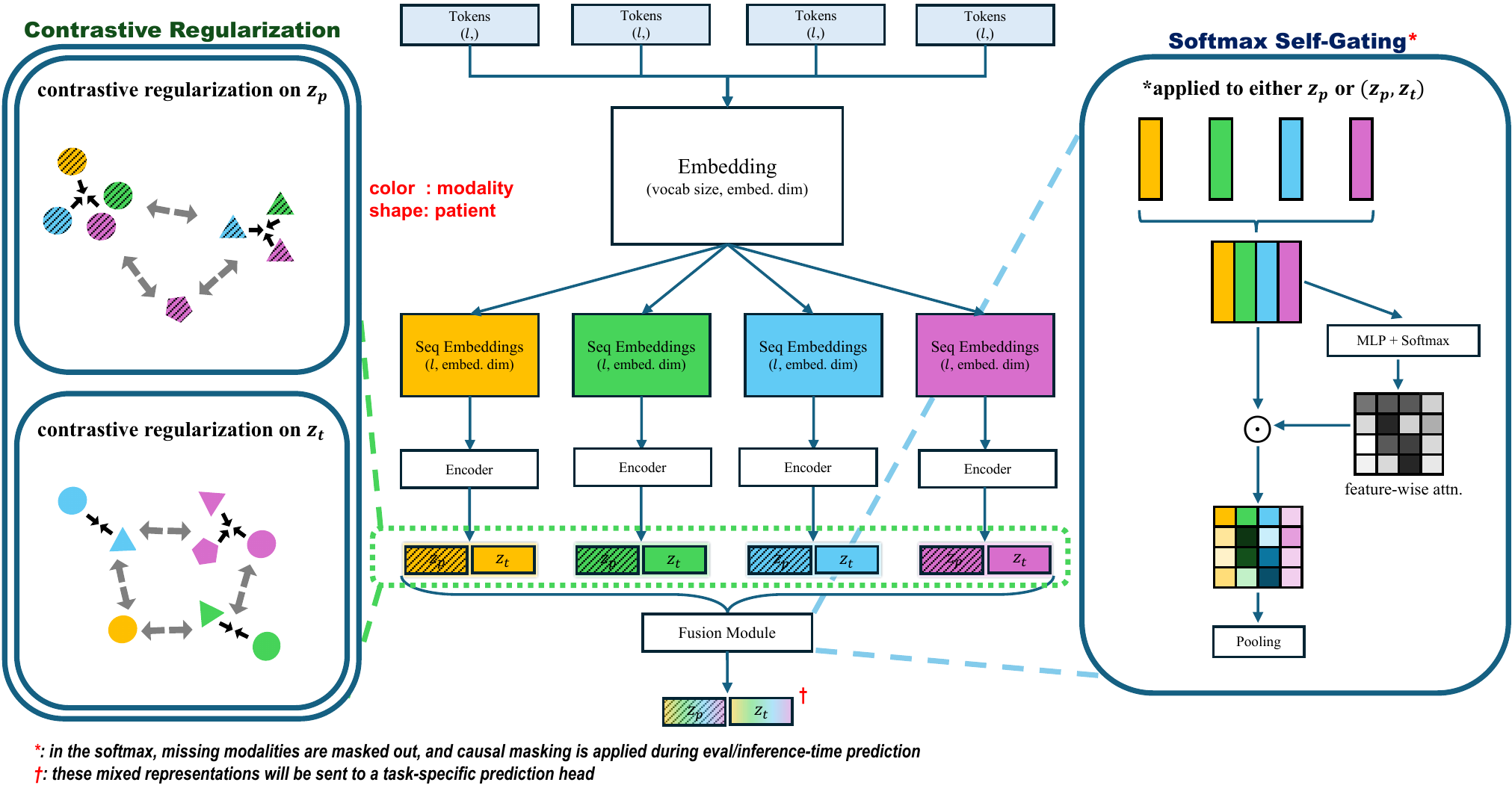}
        \caption{Multi-modal architecture with Softmax self-gating for modality fusion. Each modality is a medical code sequence of length $l$ (padded as needed), and the corresponding encoder will extract two feature vectors: $\vz_{p}$ and $\vz_{t}$. A fusion module is applied to either $\vz_{p}$ or $(\vz_{p},\vz_{t})$, depending on which serves as input in the downstream task. Here, we use an example of 3 samples with missing modalities to showcase the mechanism of the contrastive regularization. Shapes and colors correspond to unique samples and modalities, respectively.}
        \label{fig:full-model}
    \end{figure}

    We consider each time window as a separate modality. All the medical
    concepts are mapped to embeddings using a pretrained Continuous Bag-of-Word
    (CBOW) model \citep{mikolov2013efficient}. For each modality, an encoder takes
    the average pooling of the token embeddings along the sequence dimension and
    then calculates latent representation $(\vz_{p}, \vz_{t})$. Subsequently, a fusion
    module mix up the representations from different modalities. Fig. \ref{fig:full-model}
    visually illustrates working mechanism of BFF. Finally, a prediction head can
    use either $\vz_{p}$ or $(\vz_{p}, \vz_{t})$ for a downstream task.

    As demonstrated in Fig. \ref{fig:full-model}, the CR consists of two contrastive loss terms: a cross-modal loss applied to \textit{patient-specific} features ($\vz_p$), and a within-modal loss focusing on \textit{time- and modality-specific} features ($\vz_t$). Soft Nearest Neighbor (SNN) loss is used for both regularization terms \citep{frosst2019analyzing} (Sec. \ref{sec:contrastive}). Notably, these contrastive terms are additional to the loss function for the prediction task of interest. We use Softmax Self-Gating as the multi-modal fusion module to separately aggregate $\vz_{p}$ and $\vz_{t}$ from all modalities (Sec. \ref{sec:fusion}). Therefore, the overall loss function for BFF training is:
    \begin{equation}
        \cL = \cL_{\text{prediction}}+ \cL_{\text{SNN}}^{(\text{across})}+ \cL_{\text{SNN}}
        ^{(\text{within})}
    \end{equation}
    To summarize, $\cL_{\text{prediction}}$ is the objective for the prediction
    task (e.g. negative log-likelihood). $\cL_{\text{SNN}}^{(\text{across})}$
    and $\cL_{\text{SNN}}^{(\text{within})}$ are the SNN regularization terms
    for across-modal and within-modal alignment, respectively.

    \subsection{Training \& Testing of BFF and Other Approaches}

    The standard approach to developing a CPM for risk prediction at time $t$ involves using a conventional holdout validation set, where both training and testing data share a common structure. More specifically, in both datasets, the input data are drawn exclusively from historical time windows that occur \textbf{prior to} $t$.

    Recognizing that future data may embed stronger predictive cues, early risk assessment can be enhanced through a two-stage process: self-supervised forecasting pretraining followed by a task-specific fine-tuning. \citep{lyu2018improving, xue2020learning}. More specifically, (1) we pretrain a forecasting autoencoder to predict future observations based on data available before time $t$; (2) we use the pretrained encoder to extract representations of historical data, which later serve as the input features of the downstream CPM. During model testing, only information observed prior to $t$ is processed through the forecasting encoder and subsequently passed to the CPM for the prediction. See Appx.~\ref{appx:forecast} for implementation details of the forecasting approach.

    \begin{table}[h!]
        \centering
        \caption{Differences in how training and testing data are utilized to
        develop a CPM for risk assessment at time $t$ across different
        approaches}
        \label{tab:diff} \resizebox{0.8\columnwidth}{!}{
        \begin{tabular}{l|ccc}
            \toprule                   & Training Paradigm & Training Data       & Testing Data \\
            \midrule Standard Practice & one-step          & prior to $t$        & prior to $t$ \\
            Forecasting                & two-step          & throughout the time & prior to $t$ \\
            \textbf{BFF}               & one-step          & throughout the time & prior to $t$ \\
            \bottomrule
        \end{tabular}
        }
    \end{table}

    Our BFF framework, on the other hand, is more \textit{train-efficient} than the two-step forecasting procedure. The CPM will be trained using all available information across all time windows within a single step. The encoders for all time windows will be jointly optimized under the contrastive regularization. However, during the testing/evaluation phase, BFF calculates the risks at $t$ only using the information observed prior to $t$. Encoders for the future time windows will then be masked off. Tab. \ref{tab:diff} summarizes how BFF and other approaches are different in utilizing training and testing data for model development.

    \subsection{Contrastive Regularization}
    \label{sec:contrastive}

    Let the sample size be $N$ and the number of modalities be $m$. Let $\left\{\vz_{p_{ij}}\right\}_{i=1,j=1}^{N,m}$ and $\left\{\vz_{t_{ij}}\right\}_{i=1,j=1}^{N,m}$ denote the sets of patient representations and modality representations produced by the encoders, respectively, where $i$ indexes the patient and $j$ indexes the modality. Since each semantic can have multiple positive pairs, SNN loss is used for both within-modal and cross-modal contrastive learning. In our implementation of the SNN loss, we replace the $L_{2}$ distance with cosine similarity, $\text{sim}(\cdot,\cdot)$, as cosine similarity is more aligned with the established InfoNCE derivation and used more often in the mainstream contrastive learning frameworks \citep{oord2018representation, he2020momentum, chen2020simple, radford2021learning}.

    We implement a dual contrastive mechanism that jointly optimizes within-modal feature understanding and cross-modal representation alignment. In both components, similarities are encouraged to be maximized among the positively paired representations but minimized among those negatively paired ones. In the within-modal contrastive term (Eqn. \ref{eqn:snn-within}), positive pairs are made by representations belonging to the same modality (observed from the same time-window). This allows the encoders to preserve modality-specific information in $\vz_{t_{ij}}$'s. On the other hand, in the cross-modal contrastive term (Eqn. \ref{eqn:snn-across}), positive pairs are made by representations generated from the same patient. Under this regularization, $\vz_{p_{ij}}$'s are encouraged to carry patient-specific features. The contrastive temperatures $\tau_{w}$ and $\tau_{a}$ are learnable parameters. $M_i$ denotes the index set of observed modalities for patient $i$.

    \begin{equation}
        \cL_{\text{SNN}}^{(\text{within})}= \frac{1}{Nm}\sum_{i=1}^{N}\sum_{j \in M_i}
        -\log \frac{\sum_{i',j'}\mathbb{I}_{[j'=j]}\exp \left\{ \mathrm{sim}(\vz_{t_{ij}},\vz_{t_{i'j'}})/\tau_{w}\right\}}{\sum_{i',j'}\exp
        \left\{ \mathrm{sim}(\vz_{t_{ij}},\vz_{t_{i'j'}})/\tau_{w}\right\} }\label{eqn:snn-within}
    \end{equation}

    \begin{equation}
        \cL_{\text{SNN}}^{(\text{across})}= \frac{1}{Nm}\sum_{i=1}^{N}\sum_{j \in M_i}
        -\log \frac{\sum_{i',j'}\mathbb{I}_{[i'=i]}\exp \left\{ \mathrm{sim}(\vz_{p_{ij}},\vz_{p_{i'j'}})/\tau_{a}\right\}}{\sum_{i',j'}\exp
        \left\{ \mathrm{sim}(\vz_{p_{ij}},\vz_{p_{i'j'}})/\tau_{a}\right\} }\label{eqn:snn-across}
    \end{equation}

    \begin{figure}[h]
        \centering
        \begin{minipage}{0.25\textwidth}
            \centering
            \includegraphics[width=\textwidth]{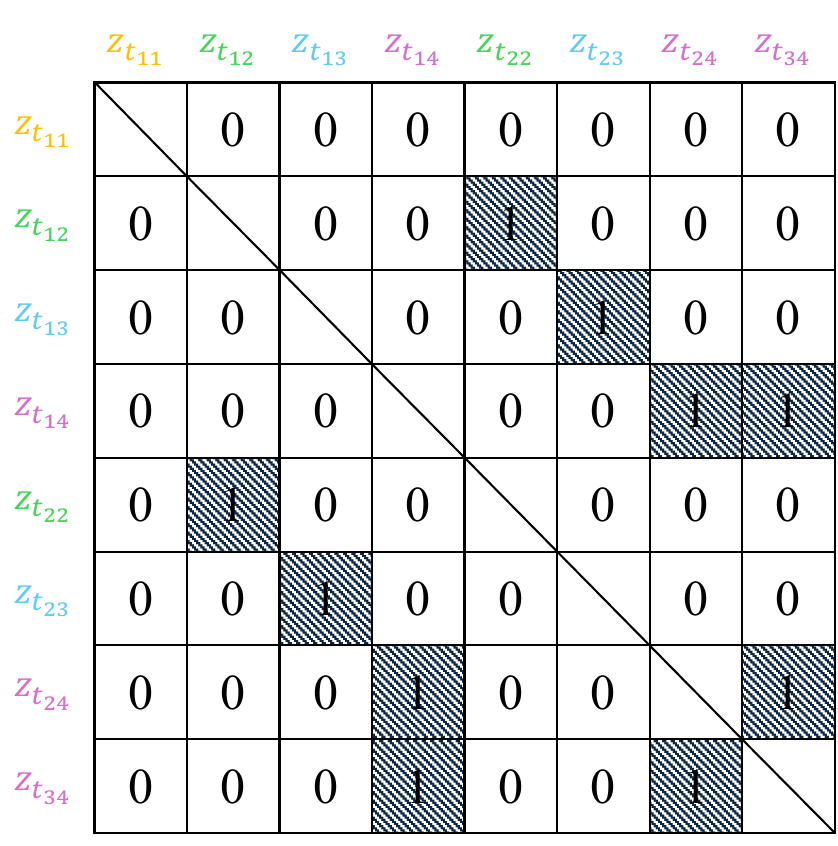}
            {\captionsetup{font=footnotesize}\caption*{(a) within-modality}}
        \end{minipage}
        \hspace{3em}
        \begin{minipage}{0.25\textwidth}
            \centering
            \includegraphics[width=\textwidth]{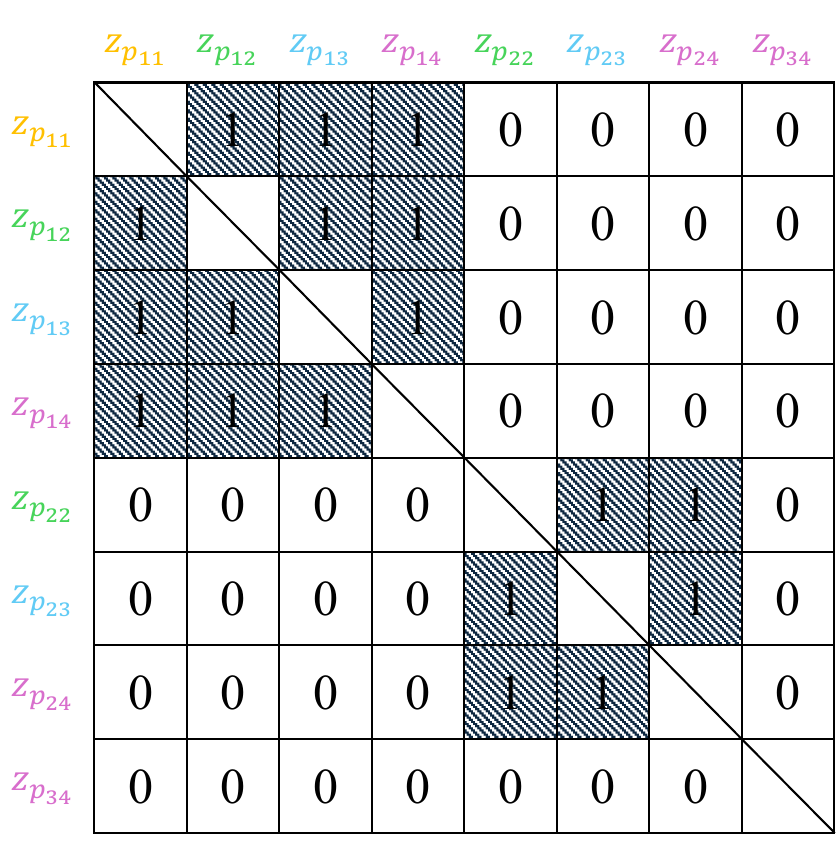}
            {\captionsetup{font=footnotesize}\caption*{(b) cross-modality}}
        \end{minipage}
        \caption{Representation pairing in the dual contrastive mechanism.}
        \label{fig:pair-mat}
    \end{figure}

    To illustrate the dual contrastive mechanism, let's consider a hypothetical scenario with 4 modalities and 3 patients, where some modalities are missing. The representations generated by the available modalities are 
    $\{\vz_{p_{11}}, \vz_{p_{12}}, \vz_{p_{13}}, \vz_{p_{14}}, \vz_{p_{22}},\vz_{p_{23}}, \vz_{p_{24}}, \vz_{p_{34}}\}$ and 
    $\{\vz_{t_{11}}, \vz_{t_{12}}, \vz_{t_{13}}, \vz_{t_{14}}, \vz_{t_{22}},\vz_{t_{23}}, \vz_{t_{24}}, \vz_{t_{34}}\}$. 
    Fig. \ref{fig:pair-mat} demonstrates how the representations are paired up, and the left panel of Fig. \ref{fig:full-model} visualizes the way CR adjusts them.

    Following our prior discussion, the representations derived from future modalities (e.g., medical codes collected from developmental Well-Child visits) are inherently more informative for predicting outcomes, as they are based on the information collected closer to the events onset. In the cross-modality alignment, these future representations serve as “soft labels” to guide the representations formed from earlier modalities. This approach is analogous to the contrastive mechanism in the CLIP framework, where natural language implicitly supervises the visual representation
    learning \citep{radford2021learning, yuan2021multimodal, lin2022multimodal}.

    \subsection{Softmax Self-Gating}
    \label{sec:fusion} We treat data from each time window as a separate modality. An effective multi-modal fusion mechanism is significant for the downstream prediction performance. Our Softmax Self-Gating for modality fusion is inspired by the SE block for computer vision tasks \citep{hu2018SEnet}. We extend the SE-block to our multi-modal setting and calculate the attention weights for each feature (e.g., embedding coordinate) from each modality. The channel-wise weights now become feature-wise attention scores of all modalities in the downstream information aggregation.

    Assume we have $m$ modalities and each has a latent embedding size of $d$,
    and for each patient, we have a multi-modal feature embedding $Z = [\vz_{1},\vz
    _{2},...,\vz_{m}]^{\top} \in \mathbb{R}^{m\times d}$. We compute the feature-wise attention weights $W = [\vw_{1},\vw_{2},...,\vw_{m}]^{\top}$ using an MLP, denoted as $g_{\phi}: \mathbb{R}^{m\times d}\to \mathbb{R}^{m\times d}$, followed by a final softmax activation that generates feature-wise attention weight distributions along the modality dimension. The gated feature matrix $Z^{*}$ is obtained by element-wise multiplying $Z$ with the gating score matrix $W$, which adaptively modulates the original features.
    \begin{equation}
        Z^{*}= Z \odot W = Z \odot \text{Softmax}(g_{\phi}(Z)) \label{eq:xstar}
    \end{equation}
    For each patient, the final representation $\vs$ is formed through feature-specific weighted sum across all modalities.
    Fig. \ref{fig:full-model} provides a visual demonstration of this process.
    \begin{equation}
        \vs = \sum_{j=1}^{m}\vz_{j}\odot \vw_{j}
    \end{equation}
    For predictions at time $t$, we mask future modalities in the softmax, setting their fusion weights to zero to prevent the model from accessing future information. This is similar to ``causal masking'' \citep{vaswani2017attention} but for downstream prediction rather than decoding. To handle missing modalities during both training and inference, we fill the missing pieces with padding tokens as non-meaningful placeholders. We then apply boolean masks to set the gating weights of these modalities to zero, ensuring the aggregation is only performed over observed modalities. Together, these masking strategies ensure that the model makes predictions based only on valid and temporally appropriate inputs.

    Beyond enabling efficient multi-modal fusion, the weight matrix $W$ also quantifies the contribution of each feature and modality to the downstream tasks during aggregation. See more details in Sec. \ref{sec:w_attn}.

    \subsection{Model Evaluation}
    We use the area under the curve (AUC) of the regular receiver operating characteristic (ROC) curve to evaluate the binary classification task performance. For the time-to-event prediction task, we use cumulative/dynamic AUC ($\text{AUC}_{\text{C/D}}$), a time-dependent extension of the original AUC, firstly introduced by \citet{heagerty2005survival}. It evaluates model's capability of differentiating \textit{cumulative cases} and \textit{dynamic controls}. Let $T_{i}$ be the true time-to-event of an arbitrary individual. For a given time $t$, cumulative cases refer to the patients who experience the event of interest by the time ($T_{i}\le t$), whereas dynamic controls are those who still remain event-free at the moment ($T_{i}> t$) \citep{heagerty2005survival, kamarudin2017time}. In this work, we use the inverse probability of censoring weighting (IPCW) estimator (Eqn.
    \ref{eq:cd-auc}) proposed by \citet{uno2007evaluating} and \citet{hung2010estimation}
    to compute the metric.
    \begin{equation}
        \widehat{\text{AUC}}_{\text{C/D}}(t) = \frac{\sum_{i=1}^{n}\sum_{j=1}^{n}\mathbb{I}(y_{j}>
        t)\mathbb{I}(y_{i}\leq t)\omega_{i}\mathbb{I}(\hat{f}(\mathbf{x}_{j}) \leq
        \hat{f}(\mathbf{x}_{i}))}{\left(\sum_{i=1}^{n}\mathbb{I}(y_{i}> t)\right)\left(\sum_{i=1}^{n}\mathbb{I}(y_{i}\leq
        t)\omega_{i}\right)}\label{eq:cd-auc}
    \end{equation}
    where $\omega_{i}= \delta_{i}/\hat{S}(y_{i})$ is the inverse of the
    probability of being censored for subject $i$ and $\delta_{i}$ is event
    indicator. The estimated survival function $\hat{S}(\cdot)$ is computed using
    Kaplan-Meier Estimator \citep{kaplan1958nonparametric} based on the training
    data.

    We use the method (Eqn. \ref{eq:intg-auc}) proposed by \citet{lambert2016summary}
    to summarize $\widehat{\text{AUC}}_{\text{C/D}}(t)$ over time. It calculates
    a weighted mean over a restricted time interval $(\tau_{1}, \tau_{2})$ by integrating
    $\widehat{\text{AUC}}_{\text{C/D}}(t)$ over the estimated survival function.

    \begin{equation}
        \overline{\text{AUC}}_{\text{C/D}}(\tau_{1}, \tau_{2}) = \frac{1}{\hat{S}(\tau_{1})
        - \hat{S}(\tau_{2})}\int_{\tau_1}^{\tau_2}\widehat{\text{AUC}}_{\text{C/D}}
        (t) \, d\hat{S}(t) \label{eq:intg-auc}
    \end{equation}

    \section{Experiment \& Data}
    \label{sec:data}
    \subsection{CBOW pre-training}
    To capture generalized representations of medical encounters for both
    children and their mothers, we trained a CBOW model on tokenized EHR codes.
    In this formulation, each code was treated as an individual token, with
    children’s codes covering ages 0–24 months and mothers’ codes spanning the
    prenatal period through newborn delivery. The data was sourced from the Duke
    Clinical Research Datamart (CRDM) \citep{hurst2021development}. Specifically,
    we used encounter data consisting of diagnosis codes, procedure codes, medication prescription and laboratory tests for all children born in 2015-2022. The data included maternal prenatal visits, birth encounters, and babies’ postnatal Well-Child visits. To prevent any potential data leakage, we excluded the downstream tasks' pre-defined testing data from the training data used for the CBOW model.

    \subsection{Downstream tasks}

    We considered two common tasks for early childhood risk assessment using
    data retrieved from three critical time windows: prenatal ($t_{\text{prenatal}}$),
    birth ($t_{\text{birth}}$), and developmental ($t_{\text{developmental}}$).
    For both tasks, EHR inputs are tokenized into medical codes and organized into
    four modalities: $m_{\text{prenatal}}$, $m^{\text{mom}}_{\text{birth}}$,
    $m^{\text{baby}}_{\text{birth}}$, and $m_{\text{developmental}}$.

    \begin{table}[h!]
        \centering
        \caption{Time windows and modalities}
        \label{tab:modalities} \resizebox{0.3\columnwidth}{!}{
        \begin{tabular}{ll}
            \toprule Time window           & Modality                                                         \\
            \midrule $t_{\text{prenatal}}$ & $m_{\text{prenatal}}$                                            \\
            $t_{\text{birth}}$             & $(m_{\text{birth}}^{\text{baby}},m_{\text{birth}}^{\text{mom}})$ \\
            $t_{\text{developmental}}$     & $m_{\text{developmental}}$                                       \\
            \bottomrule
        \end{tabular}
        }
    \end{table}

    \subsubsection{Prediction Task 1: Autism Spectrum Disorder (ASD)}
    ASD is a neuro-developmental condition that typically presents within the first years of life. The mean age of the diagnosis is around 5 years \citep{van2021age}. Children are typically screened during their 18-month well-child visit and referred for confirmatory diagnosis. Previous works has shown that early medical conditions (e.g., gastrointestinal disease) can serve as early indicators of a future ASD diagnosis \citep{alexeeff2017medical, engelhard2023predictive}. Our team has been focusing on developing automated early screening tools for ASD.

    We identified a cohort of 43,945 children who had a well-child visit at our institution between 12 and 24 months of age. We used the well-child visits closest to 18 months as the index encounter. We required two distinct medical encounters with an ICD-10 for ASD to label an ASD case \citep{guthrie2019accuracy}. Across our cohort's follow-up, 2.05\% were diagnosed with ASD. Our training and testing sizes were 35,000 and 8,945, respectively. To account for potential censoring and loss-to-follow-up, we set up our task as time-to-event prediction task. Our outcome of interest was the time to diagnosis, and we used the fixed-interval Discrete-Time Neural Network (DTNN) as the downstream prediction model \citep{lee2018deephit, hickey2024adaptive}. More details about DTNN and the corresponding loss function for prediction is provided in Appx. \ref{appx:dtnn}.

    \subsubsection{Prediction Task 2: Recurrent Acute Otitis Media (rAOM)} 
    rAOM, or recurrent ear infection, is a common medical condition in early childhood. While most children will experience an ear infection during their first years of life, approximately 20-30\% of the pediatric population will experience recurrent ear infections defined as 3 episodes within a six-month period or 4 episodes within a year period \citep{pichichero2000recurrent, kaur2017epidemiology}. Children who experience rAOM often require surgical intervention in the form of ear tube placement. We sought to develop a predictive model for the probability a first AOM develops into rAOM. We identified a cohort of 5,438 children with a first AOM. Of these, 22.0\% transitioned into rAOM. We used the same embeddings framework as above to process prenatal, birth and developmental clinical data. We split the data into training and testing sets of sizes of 4,227 and 1,211, respectively. Since our eligible cohort had reached age 4, an age when rAOM is unlikely to developed, we modeled the binary diagnosis indicator
    as the outcome.

    \section{Results}
    \subsection{Prediction Performance}

    Tab. \ref{tab:asd-rlt} and Tab. \ref{tab:raom-rlt} present model performance scores for risk assessments across the three time windows. While predictions at $t_{\text{prenatal}}$ and $t_{\text{birth}}$ are the \textbf{early risk assessments}, we include $t_{\text{developmental}}$ assessments to provide a comprehensive analysis. Same masking strategies are used for all fusion techniques. The forecasting approach does not involve any fusion and its performance at $t_{\text{development}}$ is omitted as no future data are available beyond this point. Crucially, model comparisons must be conducted \textbf{within} individual columns rather than across columns, since each column corresponds to different evaluation time points with varying testing populations. For
    example, when evaluating the performance at $t_{\text{birth}}$, the analysis is restricted to testing samples with $m_{\text{birth}}$ observed, since some
    children were not born at our institution.

    \begin{table}[h!]
        \centering
        \caption{Evaluation results of \textbf{ASD time-to-diagnosis} prediction
        task. $\overline{\text{AUC}}_{\text{C/D}}$ is the integrated measure
        across postnatal years 3-8. The prediction head $f(\cdot)$ encapsulates
        the specific representations utilized for the downstream task. CR includes
        both within-modal and across-modal contrastive losses. Performance
        results are averaged over 5 random seeds}
        \label{tab:asd-rlt} \resizebox{0.85\columnwidth}{!}{
        \begin{tabular}{llccc}
        \toprule
        \multirow{2}{*}{Method} & \multirow{2}{*}{Fusion Techniques} & \multicolumn{3}{c}{$\overline{\text{AUC}}_{\text{C/D}}$} \\
        \cmidrule(lr){3-5}
         &  & $\boldsymbol{t}_{\textbf{prenatal}}$ & $\boldsymbol{t}_{\textbf{birth}}$ & $t_{\text{developmental}}$ \\
        \midrule
        \multirow{3}{*}{Standard Practice}
         & masked mean         & 0.676 (0.004) & 0.677 (0.009) & 0.756 (0.003) \\
         & self-attention      & 0.672 (0.004) & 0.684 (0.024) & 0.785 (0.004) \\
         & softmax self-gating & 0.671 (0.011) & 0.685 (0.012) & 0.792 (0.006) \\
        \midrule
        \multirow{1}{*}{Forecasting}
         & \quad\quad\quad/ & \textbf{0.722 (0.003)} & 0.737 (0.004) & - \\
        \midrule
        \midrule
        \multirow{3}{*}{BFF w/o CR}
         & masked mean         & 0.680 (0.009) & 0.679 (0.011) & 0.760 (0.008) \\
         & self-attention      & 0.682 (0.002) & 0.682 (0.010) & 0.778 (0.009) \\
         & softmax self-gating & 0.683 (0.002) & 0.684 (0.004) & 0.783 (0.007) \\
        \midrule
        \multirow{3}{*}{\textbf{BFF + $\boldsymbol{f(\vz_p)}$}}
         & masked mean         & 0.707 (0.008)  & \textbf{0.745 (0.004)} & 0.742 (0.011) \\
         & self-attention      & 0.697 (0.011)  & 0.741 (0.005) & 0.744 (0.017) \\
         & softmax self-gating & \textbf{0.721 (0.011)}  & \textbf{0.749 (0.009)} & 0.742 (0.010)  \\
        \midrule
        \multirow{3}{*}{BFF + $f(\vz_p, \vz_t)$}
         & masked mean         & 0.706 (0.008)  & 0.733 (0.028) & 0.762 (0.003) \\
         & self-attention      & 0.690 (0.006)  & 0.733 (0.015) & 0.765 (0.009) \\
         & softmax self-gating & 0.693 (0.010)  & 0.727 (0.020) & 0.787 (0.005) \\
        \bottomrule
    \end{tabular}
        }
    \end{table}

    \begin{figure}[h!]
        \centering
        \includegraphics[width=0.8\linewidth]{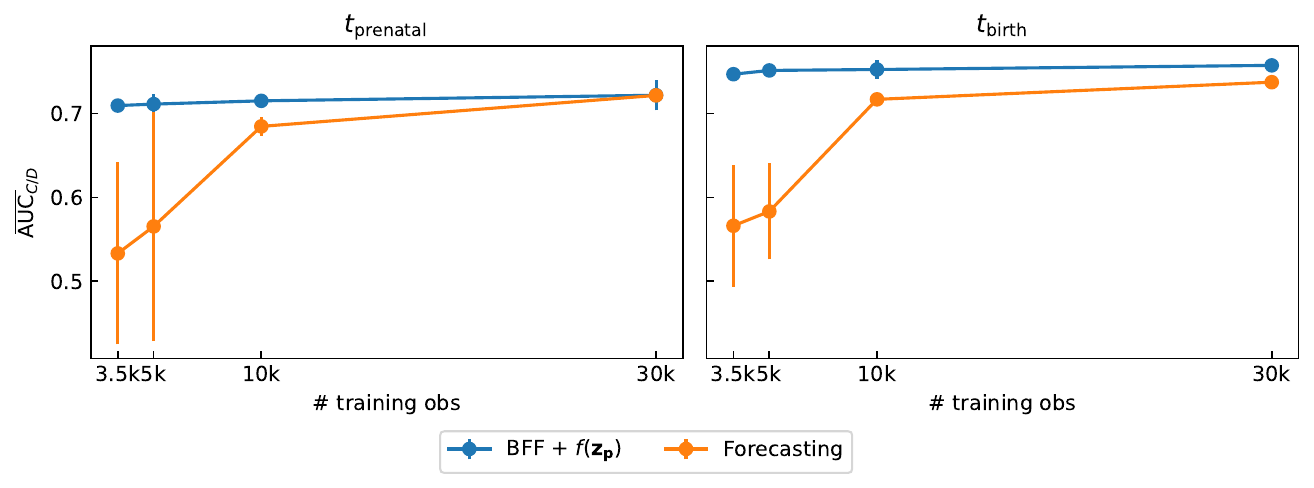}
        \caption{\textbf{BFF, as a one-step procedure, demonstrates more robust
        performance and higher data efficiency under limited training data.} The
        performance scores are averaged over five random seeds, and the error
        bars are the standard deviations.}
        \label{fig:forecast-inefficiency}
    \end{figure}

    Given the substantial size ($\sim$44k observations) of the ASD cohort, we utilize
    it to conduct experiments comparing the \textit{data efficiency} of BFF
    against the forecasting approach. We train the model with random sub-samples
    from the training data and evaluate it using the original testing data. According to Fig. \ref{fig:forecast-inefficiency}, BFF demonstrates superior data efficiency compared to the forecasting approach. Moreover, BFF achieves robust performance with significantly less amount of data, while the forecasting approach requires much more data to reach comparable levels. The forecasting method demonstrates inferior data efficiency and poor generalizability (i.e. unstable performance) under limited training data, as evidenced by low performance scores and high variability using small random training subsets. The model fails to extract robust patterns from the scarce data and becomes overly dependent on the specific training instances selected. Therefore, when applied to downstream tasks with small datasets, such as our rAOM dataset containing merely $\sim$4k observations, the forecasting approach is expected to have significantly degraded
    performance.

    \begin{table}[h!]
        \centering
        \caption{Evaluation results of \textbf{rAOM binary} prediction task. The
        prediction head $f(\cdot)$ encapsulates the specific representations
        utilized for the downstream task. Performance results are averaged over 5
        random seeds.}
        \resizebox{0.85\columnwidth}{!}{
        \begin{tabular}{llccc}
        \toprule
        \multirow{2}{*}{Method} & \multirow{2}{*}{Fusion Techniques} & \multicolumn{3}{c}{AUC} \\
        \cmidrule(lr){3-5}
         &  & $\boldsymbol{t}_{\textbf{prenatal}}$ & $\boldsymbol{t}_{\textbf{birth}}$ & $t_{\text{developmental}}$ \\
        \midrule
        \multirow{3}{*}{Standard Practice}
         & masked mean         & 0.613 (0.002) & 0.612 (0.005) & 0.871 (0.003) \\
         & self-attention      & 0.604 (0.002) & 0.610 (0.003) & 0.872 (0.001) \\
         & softmax self-gating & 0.600 (0.012) & 0.613 (0.007) & 0.870 (0.002) \\
        \midrule
        \multirow{1}{*}{Forecasting}
         & \quad\quad\quad/ & 0.519 (0.042) & 0.505 (0.010) & - \\
        \midrule
        \midrule
        \multirow{3}{*}{BFF w/o CR}
         & masked mean         & 0.580 (0.018) & 0.597 (0.003) & 0.868 (0.001) \\
         & self-attention      & 0.600 (0.006) & 0.620 (0.005) & 0.868 (0.003) \\
         & softmax self-gating & 0.573 (0.001) & 0.583 (0.003) & 0.871 (0.002) \\
        \midrule
        \multirow{3}{*}{\textbf{BFF + $\boldsymbol{f(\vz_p)}$}}
         & masked mean         & \textbf{0.628 (0.010)} & \textbf{0.634 (0.012)} & 0.657 (0.002) \\
         & self-attention      & 0.609 (0.014) & 0.622 (0.007) & 0.649 (0.002) \\
         & softmax self-gating & \textbf{0.623 (0.006)} & \textbf{0.641 (0.004)} & 0.649 (0.021) \\
        \midrule
        \multirow{3}{*}{BFF + $f(\vz_p, \vz_t)$}
         & masked mean         & 0.600 (0.013) & 0.608 (0.015) & 0.856 (0.002) \\
         & self-attention      & 0.611 (0.015) & 0.607 (0.016) & 0.857 (0.003) \\
         & softmax self-gating & 0.605 (0.010) & 0.628 (0.009) & 0.852 (0.012) \\
        \bottomrule
    \end{tabular}
        } \label{tab:raom-rlt}
    \end{table}

    The BFF framework demonstrates significant improvement in early risk assessment performance. The contrastive regularization terms emerge as the primary driver of the enhancements, particularly through the improved learning of patient-specific features, $\vz_{p}$. One interesting observation is that using the time-/modality-sensitive features, $\vz_{t}$, can be counter-effective for the predictions at $t_{\text{prenatal}}$ and $t_{\text{birth}}$. This implies that the $\vz_{t}$ from the earlier modalities can obscure the true signal. Nevertheless, the $\vz_{t}$ from $m_{\text{developmental}}$ is highly predictive of the outcome. The result also empirically validates that risk assessments conducted later in time are generally more accurate since the data observed later in time is closer and more correlated
    to the potential outcome of interest.

    \subsection{Modality Attention during BFF's Training Stage}
    \label{sec:w_attn}

    \begin{figure}[!h]
        \centering
        \begin{minipage}{0.32\textwidth}
            \centering
            \includegraphics[width=\textwidth]{
                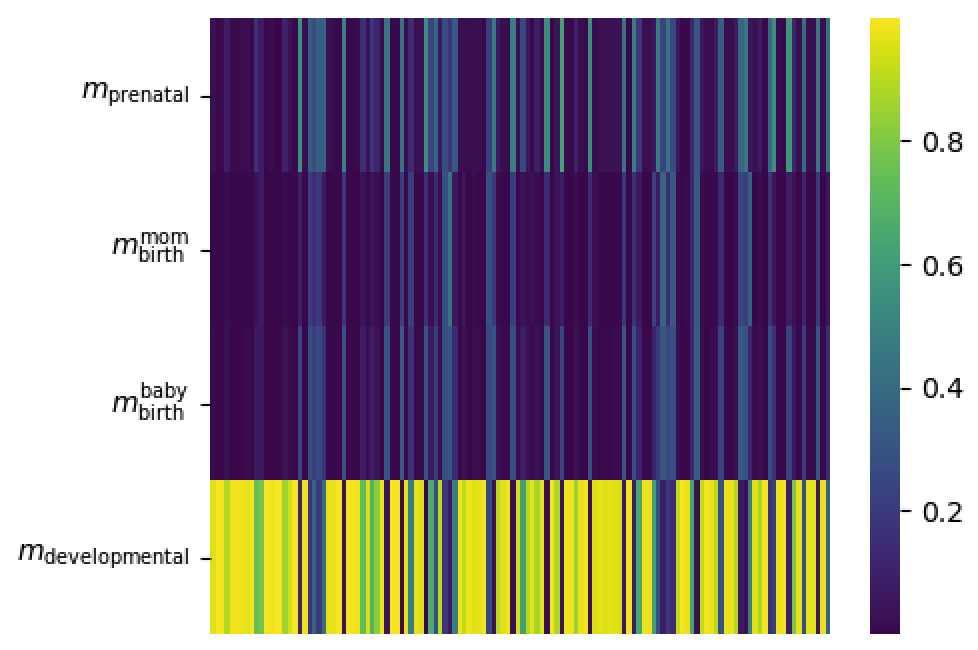
            }
        \end{minipage}
        \hfill
        \begin{minipage}{0.32\textwidth}
            \centering
            \includegraphics[width=\textwidth]{
                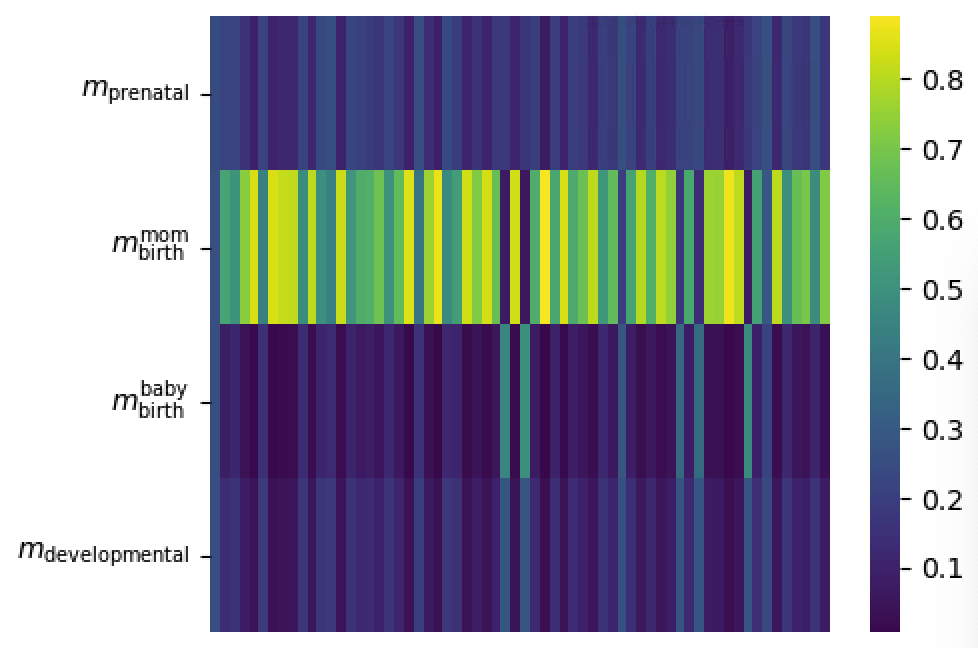
            }
        \end{minipage}
        \hfill
        \begin{minipage}{0.32\textwidth}
            \centering
            \includegraphics[width=\textwidth]{
                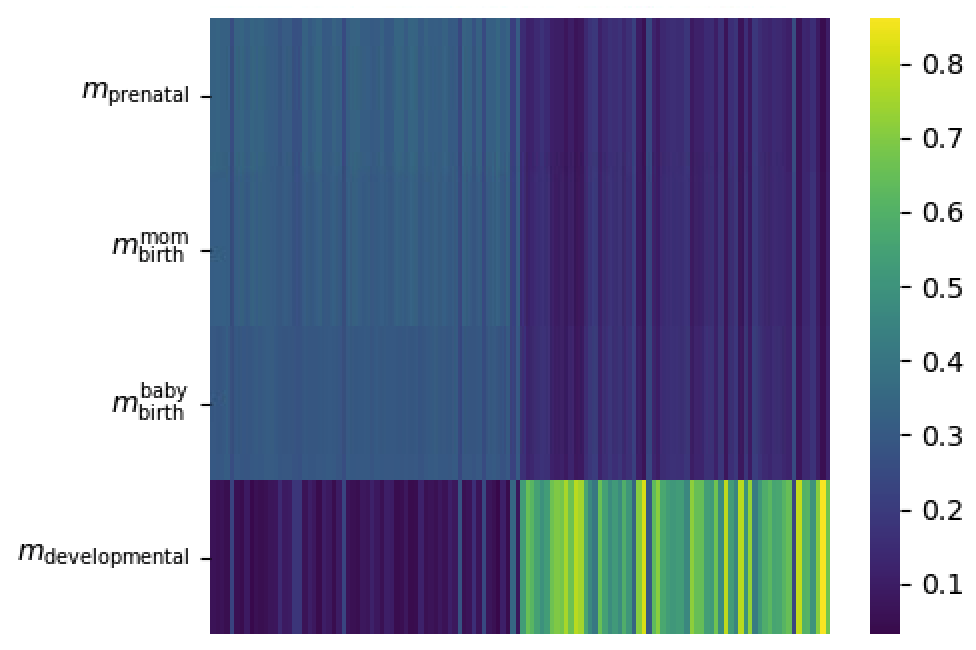
            }
        \end{minipage}

        \vspace{1em} 

        \begin{minipage}{0.32\textwidth}
            \centering
            \includegraphics[width=\textwidth]{
                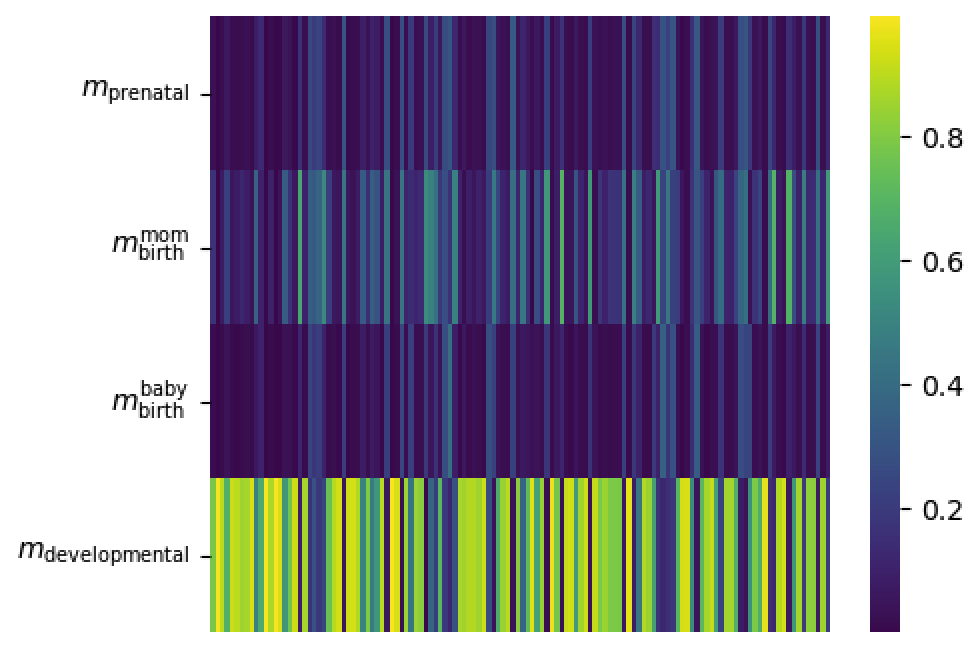
            }
            {\captionsetup{font=footnotesize}\caption*{(a) BFF w/o CR}}
        \end{minipage}
        \hfill
        \begin{minipage}{0.32\textwidth}
            \centering
            \includegraphics[width=\textwidth]{
                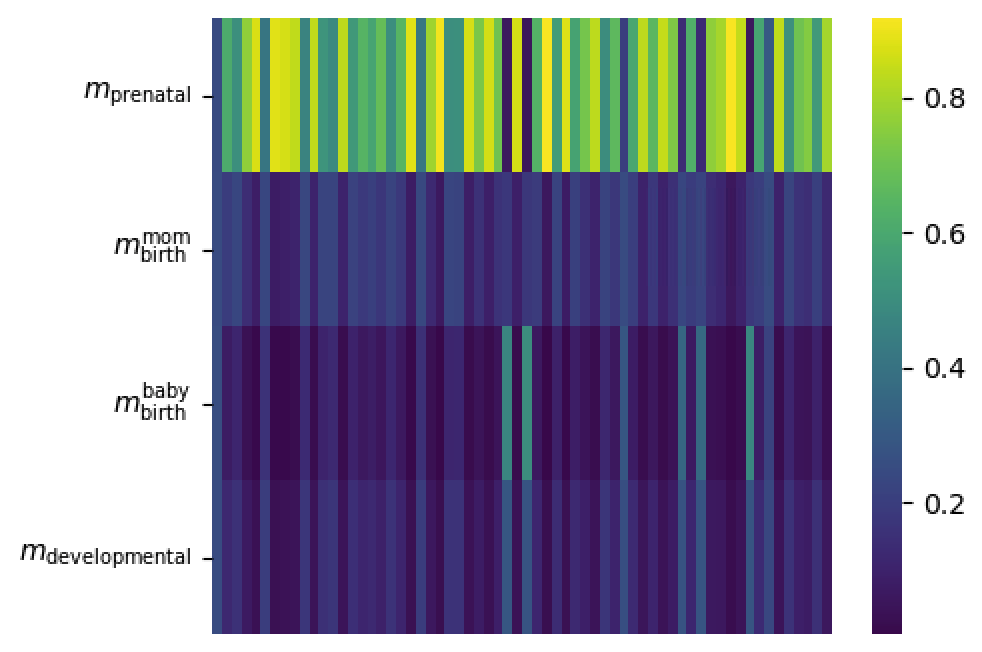
            }
            {\captionsetup{font=footnotesize}\caption*{(b) BFF + $f(\vz_{p})$}}
        \end{minipage}
        \hfill
        \begin{minipage}{0.32\textwidth}
            \centering
            \includegraphics[width=\textwidth]{
                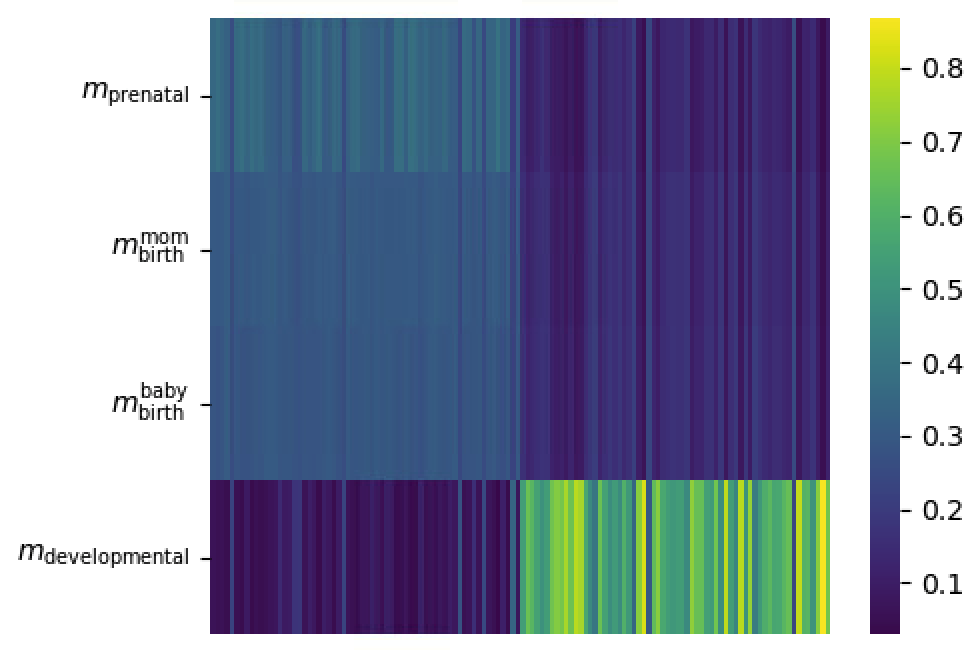
            }
            {\captionsetup{font=footnotesize}\caption*{(c) BFF + $f(\vz_{p}, \vz_{t})$}}
        \end{minipage}

        \caption{Feature-wise attention scores from the Softmax Self-Gating in the ASD time-to-diagnosis prediction. Each row corresponds to a patient, and the weight matrices, visualized by the heat maps, are calculated for the patient in different model setup. The y-axis indicates the modality while the x-axis indicates the feature vector coordinates. There is no differentiation between $\vz_{p}$ and $\vz_{t}$ without CR.}
        \label{fig:attn-self-gating}
    \end{figure}

    The commonly used self-attention computes pair-wise attention across a sequence of modality. These scores represent how much attention each modality should allocate to others. While these weights produce contextualized representation features, it is challenging to translate them to feature importance measures. Appx. \ref{sec:self-attn} provides visualizations of the multi-modal self-attention scores. However, Softmax Self-Gating calculates a input-dependent weight for each coordinate in the embeddings, which can be directly interpreted as feature attribution to the downstream task. It offers an explainable illustration of which modalities/time windows are more emphasized during the learning. More importantly, it explains why BFF improves from a standard practice in early risk assessment.

    Fig. \ref{fig:attn-self-gating} presents the Softmax self-gating weight matrices employed by BFF for feature modulation during training, illustrated using two examples under different configurations. Evidenced by (a), when all modalities are used during the training without CR, $m_{\text{developmental}}$ disproportionately dominates the calculation. The model naturally prioritizes $m_{\text{developmental}}$'s representation due to its high predictability to the outcome. Therefore, it underutilizes the early-stage modalities, and insufficiently optimizes the corresponding encoders. With the contrastive regularization in BFF, the encoders for the early modalities utilize the representations from later modalities as soft supervision. In (b), we can observe that model is regularized to direct greater attention to $m_{\text{prenatal}}$ an $m_{\text{birth}}^{\text{mom}}$ when the downstream prediction head only takes $\vz_{p}$, thereby enhancing the early risk assessment. Additionally, (c) indicates that when both $\vz_{p}$ and $\vz_{t}$ from all modalities are used in the downstream prediction, the prediction head would prioritize the modality-specific features $\vz_{t}$ from the developmental modality and pay less attention to $\vz_{p}$ from the early-stage modalities. Consequently, the encoders may not fully leverage the benefits of the borrowing from the future information.

\begin{figure}[ht]
        \centering
        \begin{minipage}{0.75\linewidth}
            \centering
            \includegraphics[width=\linewidth]{
                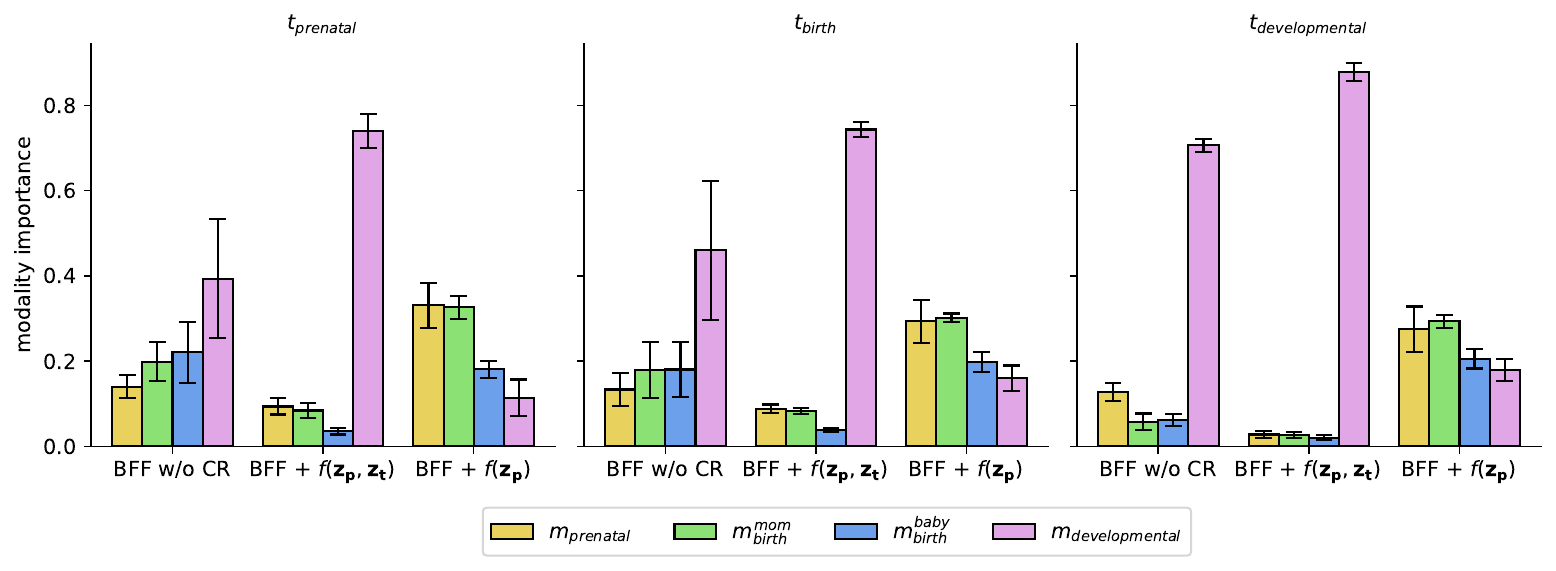
            }
            \vspace{-0.75cm}
            {\captionsetup{font=footnotesize}\caption*{(a) ASD}}
        \end{minipage}

        \vspace{0.25cm}

        \begin{minipage}{0.75\linewidth}
            \centering
            \includegraphics[width=\linewidth]{
                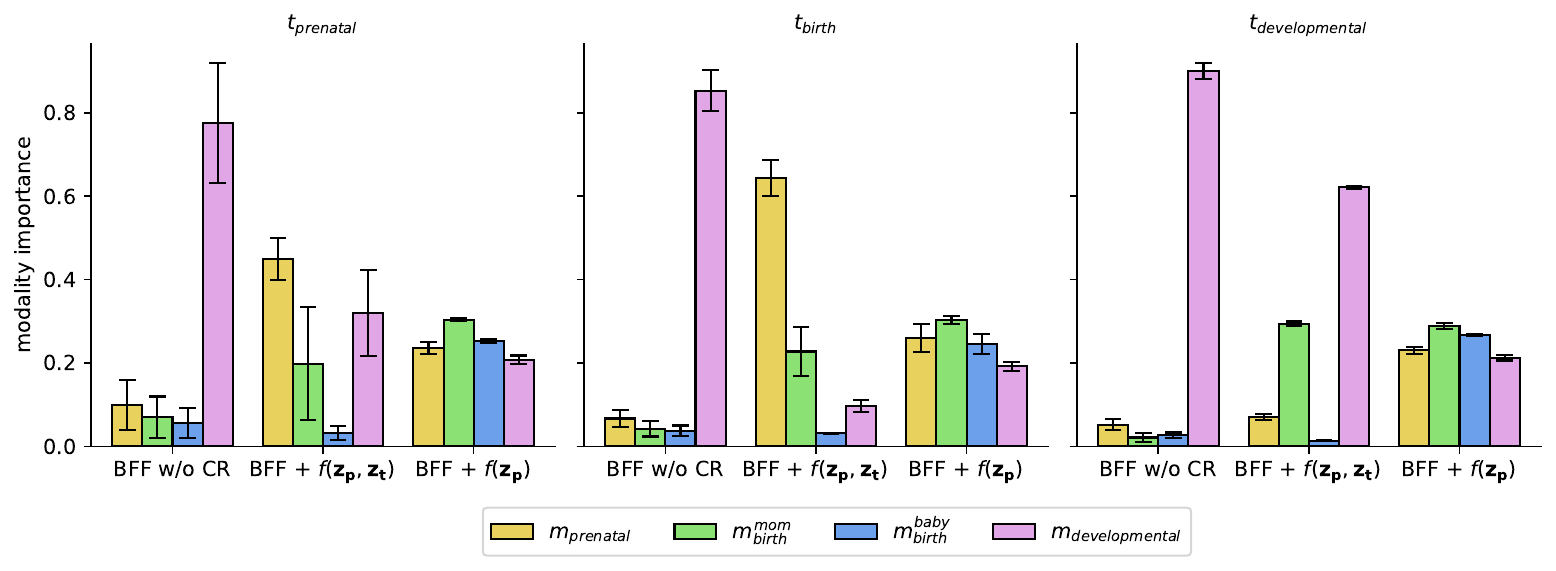
            }
            \vspace{-0.75cm}
            {\captionsetup{font=footnotesize}\caption*{(b) rAOM}}
        \end{minipage}

        \caption{Modality importance for models optimized at different evaluation time points (averaged over five random seeds). Although all models are trained on the same data spanning all time windows, the choice of evaluation time influences early stopping and checkpoint selection, which in turn affects the learned importance of each modality.}
        \label{fig:ig}
    \end{figure}

    We also corroborate these findings with the modality-level Integrated
    Gradients (IG) analysis \citep{sundararajan2017axiomatic}, shown in Fig. \ref{fig:ig}. The computing procedure details are provided in Appx. \ref{appx:ig-percent}. Contrastive regularization and latent feature selection notably influence each modality's contribution to the downstream task. In the ``BFF + $f(\vz_{p})$'' configuration, earlier modalities receive improved emphasis while the developmental modality is de-emphasized. This analysis also provides additional insight into the compromised performance at $t_{\text{development}}$ of ``BFF + $f(\vz_{p})$'': when it comes to the risk assessment at $t_{\text{development}}$, the model still over-rely on the earlier time windows instead of focusing on the more informative developmental modality, $m_{\text{developmental}}$. Moreover, the high attributions from $m_{\text{prenatal}}$ and $m_{\text{birth}}^{\text{mom}}$ in ``BFF + $f(\vz_{p}, \vz_{t})$'' indicate these modalities have strong signals for the prediction of rAOM. Recent studies have found an association between maternal antibiotic use and early childhood ear infections \citep{pedersen2017antibiotics, hu2021association}.

    \section{Discussion}

    Inspired by previous works in contrastive learning \citep{oord2018representation, tian2020contrastive, radford2021learning},
    we introduce BFF, a novel framework that ``borrows'' the future data as soft and implicit supervision to enhance the learning of early information . Compared to the standard practice and the forecasting approach \citep{lyu2018improving, xue2020learning}, BFF achieves significant improvements in early risk assessment while maintaining high training and data efficiency. As an alternative framework, it algorithmically provides more accurate preventative decision support. Our results also shows that the contrastive regularization is a the crucial component of this framework.
    
    Additionally, we demonstrate that the Softmax Self-Gating mechanism not only facilitates effective multi-modal fusion but also reveals how contrastive regularization shapes representation learning in the BFF framework. The visualizations of feature-wise attention scores provide evidence to explain the enhancements in early risk assessments.

    Although missing data is a common challenge in developing clinical
    prediction models in real-world settings, handling it appropriately and effectively remains technically demanding. Commonly used imputation-based methods are typically designed for structured tabular data \citep{van2011mice, stekhoven2012missforest}. They define missingness at the feature level. We process EHR data into sequences of tokenized medical events following an NLP-style data structure, where missingness manifests at entire modality/time-window levels rather than individual feature values. Consequently, traditional missing data imputation methods are unsuitable for this context. Our approach leverages masking during multi-modal fusion to handle the missing modality problem. It is both simple and entirely data-driven, requiring no distributional assumptions about missingness patterns. Nevertheless, theoretically-grounded methodologies may yield better representation learning and predictive performance \citep{ma2021smil, lin2023missmodal, yao2024drfuse}. It is interesting to systematically investigate how various modality-level missing data handling approaches would impact model performances in our future work.

    \paragraph{Limitations}
    We demonstrate that BFF can enhance early risk assessments, enabling more preventive care for children. However, this comes with a trade-off: while BFF improves model performance at earlier stages, it performs worse than the standard practice at $t_{\text{developmental}}$. When solely relying on $\vz_{p}$ for downstream tasks, the model performance is notably lower at $t_{\text{developmental}}$ because the representation of this time window cannot benefit from ``borrowed'' information via cross-modal alignment and may even be adversely affected by it. The regularization term encourages $\vz _{p}$ to only capture patient-specific. To incorporate time-specific features, we use $\vz_{t}$, regularized by the within-modal term, together with $\vz_{p}$ as the input to the downstream head. However, the combined representation still slightly underperforms the baseline at $t_{\text{developmental}}$, suggesting that further work is needed to refine the CR for more effective representation learning.

    Moreover, BFF may not be generalizable to time series of continuous measures, where measurements can be highly correlated across time. Therefore, the representation learning at an earlier time point may not benefit from BFF through the contrastive regularization. Further investigation is needed to identify when BFF provides the greatest advantage and under what conditions its effectiveness may be constrained.

    Currently, in BFF, we train different models for each evaluation time. In a future work, we aim to develop a single unified model optimized across the all time windows. Rather than treating the entire developmental period as a single modality, we are also interested in increasing the temporal granularity to enable risk assessments at each well-child visit. Additionally, replacing the CBOW encoder with a Transformer-based encoder may improve modeling capacity and contextual understanding.

    \acks{We thank the anonymous reviewers for their valuable insights and constructive feedback. Special appreciation goes to Jillian Hurst, Congwen Zhao, Abby Scheer for the careful preparation and setup of the cohort data. This research was supported by NIH grants NICHD P50 HD0N3074 and NIAID KO1 AI73398. Matthew Engelhard was supported by NIMH K01 MH127409. This project was completed during the Duke AI Health Data Science Fellowship Program. Health Data Science at Duke is supported by the National Center for Advancing Translational Sciences (NCATS), National Institutes of Health, through Grant Award Number UL1 TR002553. The Duke AI Health Data Science Fellowship Program is supported by the above NCATS grant, the Duke Department of Biostatistics \& Bioinformatics, and Duke AI Health. The Duke Protected Analytics Computing Environment (PACE) program is supported by the above grant and by Duke University Health System. The content of this publication is solely the responsibility of the authors and does not necessarily represent the official views of the NIH.}


\begin{thebibliography}{50}
\providecommand{\natexlab}[1]{#1}
\providecommand{\url}[1]{\texttt{#1}}
\expandafter\ifx\csname urlstyle\endcsname\relax
  \providecommand{\doi}[1]{doi: #1}\else
  \providecommand{\doi}{doi: \begingroup \urlstyle{rm}\Url}\fi

\bibitem[Alexeeff et~al.(2017)Alexeeff, Yau, Qian, Davignon, Lynch, Crawford, Davis, and Croen]{alexeeff2017medical}
Stacey~E Alexeeff, Vincent Yau, Yinge Qian, Meghan Davignon, Frances Lynch, Phillip Crawford, Robert Davis, and Lisa~A Croen.
\newblock Medical conditions in the first years of life associated with future diagnosis of asd in children.
\newblock \emph{Journal of autism and developmental disorders}, 47:\penalty0 2067--2079, 2017.

\bibitem[Beijers et~al.(2010)Beijers, Jansen, Riksen-Walraven, and de~Weerth]{beijers2010maternal}
Roseriet Beijers, Jarno Jansen, Marianne Riksen-Walraven, and Carolina de~Weerth.
\newblock Maternal prenatal anxiety and stress predict infant illnesses and health complaints.
\newblock \emph{Pediatrics}, 126\penalty0 (2):\penalty0 e401--e409, 2010.

\bibitem[Boag(1949)]{boag1949maximum}
John~W Boag.
\newblock Maximum likelihood estimates of the proportion of patients cured by cancer therapy.
\newblock \emph{Journal of the Royal Statistical Society. Series B (Methodological)}, 11\penalty0 (1):\penalty0 15--53, 1949.

\bibitem[Chen et~al.(2020)Chen, Kornblith, Norouzi, and Hinton]{chen2020simple}
Ting Chen, Simon Kornblith, Mohammad Norouzi, and Geoffrey Hinton.
\newblock A simple framework for contrastive learning of visual representations.
\newblock In \emph{International conference on machine learning}, pages 1597--1607. PmLR, 2020.

\bibitem[Cho et~al.(2014)Cho, Van~Merri{\"e}nboer, Gulcehre, Bahdanau, Bougares, Schwenk, and Bengio]{cho2014learning}
Kyunghyun Cho, Bart Van~Merri{\"e}nboer, Caglar Gulcehre, Dzmitry Bahdanau, Fethi Bougares, Holger Schwenk, and Yoshua Bengio.
\newblock Learning phrase representations using rnn encoder-decoder for statistical machine translation.
\newblock \emph{arXiv preprint arXiv:1406.1078}, 2014.

\bibitem[Dauphin et~al.(2017)Dauphin, Fan, Auli, and Grangier]{dauphin2017language}
Yann~N Dauphin, Angela Fan, Michael Auli, and David Grangier.
\newblock Language modeling with gated convolutional networks.
\newblock In \emph{International conference on machine learning}, pages 933--941. PMLR, 2017.

\bibitem[Engelhard et~al.(2023)Engelhard, Henao, Berchuck, Chen, Eichner, Herkert, Kollins, Olson, Perrin, Rogers, et~al.]{engelhard2023predictive}
Matthew~M Engelhard, Ricardo Henao, Samuel~I Berchuck, Junya Chen, Brian Eichner, Darby Herkert, Scott~H Kollins, Andrew Olson, Eliana~M Perrin, Ursula Rogers, et~al.
\newblock Predictive value of early autism detection models based on electronic health record data collected before age 1 year.
\newblock \emph{JAMA network open}, 6\penalty0 (2):\penalty0 e2254303--e2254303, 2023.

\bibitem[Farewell(1982)]{farewell1982use}
Vern~T Farewell.
\newblock The use of mixture models for the analysis of survival data with long-term survivors.
\newblock \emph{Biometrics}, pages 1041--1046, 1982.

\bibitem[Frosst et~al.(2019)Frosst, Papernot, and Hinton]{frosst2019analyzing}
Nicholas Frosst, Nicolas Papernot, and Geoffrey Hinton.
\newblock Analyzing and improving representations with the soft nearest neighbor loss.
\newblock In \emph{International conference on machine learning}, pages 2012--2020. PMLR, 2019.

\bibitem[Gao et~al.(2021)Gao, Yao, and Chen]{gao2021simcse}
Tianyu Gao, Xingcheng Yao, and Danqi Chen.
\newblock Simcse: Simple contrastive learning of sentence embeddings.
\newblock \emph{arXiv preprint arXiv:2104.08821}, 2021.

\bibitem[Guthrie et~al.(2019)Guthrie, Wallis, Bennett, Brooks, Dudley, Gerdes, Pandey, Levy, Schultz, and Miller]{guthrie2019accuracy}
Whitney Guthrie, Kate Wallis, Amanda Bennett, Elizabeth Brooks, Jesse Dudley, Marsha Gerdes, Juhi Pandey, Susan~E Levy, Robert~T Schultz, and Judith~S Miller.
\newblock Accuracy of autism screening in a large pediatric network.
\newblock \emph{Pediatrics}, 144\penalty0 (4), 2019.

\bibitem[Hager et~al.(2023)Hager, Menten, and Rueckert]{hager2023best}
Paul Hager, Martin~J Menten, and Daniel Rueckert.
\newblock Best of both worlds: Multimodal contrastive learning with tabular and imaging data.
\newblock In \emph{Proceedings of the IEEE/CVF Conference on Computer Vision and Pattern Recognition}, pages 23924--23935, 2023.

\bibitem[He et~al.(2020)He, Fan, Wu, Xie, and Girshick]{he2020momentum}
Kaiming He, Haoqi Fan, Yuxin Wu, Saining Xie, and Ross Girshick.
\newblock Momentum contrast for unsupervised visual representation learning.
\newblock In \emph{Proceedings of the IEEE/CVF conference on computer vision and pattern recognition}, pages 9729--9738, 2020.

\bibitem[Heagerty and Zheng(2005)]{heagerty2005survival}
Patrick~J Heagerty and Yingye Zheng.
\newblock Survival model predictive accuracy and roc curves.
\newblock \emph{Biometrics}, 61\penalty0 (1):\penalty0 92--105, 2005.

\bibitem[Hickey et~al.(2024)Hickey, Henao, Wojdyla, Pencina, and Engelhard]{hickey2024adaptive}
Jimmy Hickey, Ricardo Henao, Daniel Wojdyla, Michael Pencina, and Matthew Engelhard.
\newblock Adaptive discretization for event prediction (adept).
\newblock In \emph{International Conference on Artificial Intelligence and Statistics}, pages 1351--1359. PMLR, 2024.

\bibitem[Hochreiter and Schmidhuber(1997)]{hochreiter1997long}
Sepp Hochreiter and J{\"u}rgen Schmidhuber.
\newblock Long short-term memory.
\newblock \emph{Neural computation}, 9\penalty0 (8):\penalty0 1735--1780, 1997.

\bibitem[Hu et~al.(2018)Hu, Shen, and Sun]{hu2018SEnet}
Jie Hu, Li~Shen, and Gang Sun.
\newblock Squeeze-and-excitation networks.
\newblock In \emph{Proceedings of the IEEE conference on computer vision and pattern recognition}, pages 7132--7141, 2018.

\bibitem[Hu et~al.(2021)Hu, Wang, Harwell, and Wake]{hu2021association}
Yanhong~J Hu, Jing Wang, Joseph~I Harwell, and Melissa Wake.
\newblock Association of in utero antibiotic exposure on childhood ear infection trajectories: Results from a national birth cohort study.
\newblock \emph{Journal of paediatrics and child health}, 57\penalty0 (7):\penalty0 1023--1030, 2021.

\bibitem[Hung and Chiang(2010)]{hung2010estimation}
Hung Hung and Chin-Tsang Chiang.
\newblock Estimation methods for time-dependent auc models with survival data.
\newblock \emph{Canadian Journal of Statistics}, 38\penalty0 (1):\penalty0 8--26, 2010.

\bibitem[Hurst et~al.(2021)Hurst, Liu, Maxson, Permar, Boulware, and Goldstein]{hurst2021development}
Jillian~H Hurst, Yaxing Liu, Pamela~J Maxson, Sallie~R Permar, L~Ebony Boulware, and Benjamin~A Goldstein.
\newblock Development of an electronic health records datamart to support clinical and population health research.
\newblock \emph{Journal of clinical and translational science}, 5\penalty0 (1):\penalty0 e13, 2021.

\bibitem[Kamarudin et~al.(2017)Kamarudin, Cox, and Kolamunnage-Dona]{kamarudin2017time}
Adina~Najwa Kamarudin, Trevor Cox, and Ruwanthi Kolamunnage-Dona.
\newblock Time-dependent roc curve analysis in medical research: current methods and applications.
\newblock \emph{BMC medical research methodology}, 17:\penalty0 1--19, 2017.

\bibitem[Kaplan and Meier(1958)]{kaplan1958nonparametric}
Edward~L Kaplan and Paul Meier.
\newblock Nonparametric estimation from incomplete observations.
\newblock \emph{Journal of the American statistical association}, 53\penalty0 (282):\penalty0 457--481, 1958.

\bibitem[Kaur et~al.(2017)Kaur, Morris, and Pichichero]{kaur2017epidemiology}
Ravinder Kaur, Matthew Morris, and Michael~E Pichichero.
\newblock Epidemiology of acute otitis media in the postpneumococcal conjugate vaccine era.
\newblock \emph{Pediatrics}, 140\penalty0 (3), 2017.

\bibitem[Kong et~al.(2024)Kong, Tao, Xiao, Xiong, Wei, and Cai]{kong2024predicting}
Deming Kong, Ye~Tao, Haiyan Xiao, Huini Xiong, Weizhong Wei, and Miao Cai.
\newblock Predicting preterm birth using auto-ml frameworks: a large observational study using electronic inpatient discharge data.
\newblock \emph{Frontiers in Pediatrics}, 12:\penalty0 1330420, 2024.

\bibitem[Lambert and Chevret(2016)]{lambert2016summary}
J{\'e}r{\^o}me Lambert and Sylvie Chevret.
\newblock Summary measure of discrimination in survival models based on cumulative/dynamic time-dependent roc curves.
\newblock \emph{Statistical methods in medical research}, 25\penalty0 (5):\penalty0 2088--2102, 2016.

\bibitem[Lee et~al.(2018)Lee, Zame, Yoon, and Van Der~Schaar]{lee2018deephit}
Changhee Lee, William Zame, Jinsung Yoon, and Mihaela Van Der~Schaar.
\newblock Deephit: A deep learning approach to survival analysis with competing risks.
\newblock In \emph{Proceedings of the AAAI conference on artificial intelligence}, volume~32, 2018.

\bibitem[Lee et~al.(2023)Lee, Park, and Lee]{lee2023soft}
Seunghan Lee, Taeyoung Park, and Kibok Lee.
\newblock Soft contrastive learning for time series.
\newblock \emph{arXiv preprint arXiv:2312.16424}, 2023.

\bibitem[Lin and Hu(2022)]{lin2022multimodal}
Ronghao Lin and Haifeng Hu.
\newblock Multimodal contrastive learning via uni-modal coding and cross-modal prediction for multimodal sentiment analysis.
\newblock In \emph{Findings of the Association for Computational Linguistics: EMNLP 2022}, pages 511--523, 2022.

\bibitem[Lin and Hu(2023)]{lin2023missmodal}
Ronghao Lin and Haifeng Hu.
\newblock Missmodal: Increasing robustness to missing modality in multimodal sentiment analysis.
\newblock \emph{Transactions of the Association for Computational Linguistics}, 11, 2023.

\bibitem[Lipkin et~al.(2020)Lipkin, Macias, Norwood, Brei, Davidson, Davis, Ellerbeck, Houtrow, Hyman, Kuo, et~al.]{lipkin2020promoting}
Paul~H Lipkin, Michelle~M Macias, Kenneth~W Norwood, Timothy~J Brei, Lynn~F Davidson, Beth~Ellen Davis, Kathryn~A Ellerbeck, Amy~J Houtrow, Susan~L Hyman, Dennis~Z Kuo, et~al.
\newblock Promoting optimal development: identifying infants and young children with developmental disorders through developmental surveillance and screening.
\newblock \emph{Pediatrics}, 145\penalty0 (1), 2020.

\bibitem[Lyu et~al.(2018)Lyu, Hueser, Hyland, Zerveas, and Raetsch]{lyu2018improving}
Xinrui Lyu, Matthias Hueser, Stephanie~L Hyland, George Zerveas, and Gunnar Raetsch.
\newblock Improving clinical predictions through unsupervised time series representation learning.
\newblock \emph{arXiv preprint arXiv:1812.00490}, 2018.

\bibitem[Ma et~al.(2021)Ma, Ren, Zhao, Tulyakov, Wu, and Peng]{ma2021smil}
Mengmeng Ma, Jian Ren, Long Zhao, Sergey Tulyakov, Cathy Wu, and Xi~Peng.
\newblock Smil: Multimodal learning with severely missing modality.
\newblock In \emph{Proceedings of the AAAI Conference on Artificial Intelligence}, volume~35, pages 2302--2310, 2021.

\bibitem[Mikolov et~al.(2013)Mikolov, Chen, Corrado, and Dean]{mikolov2013efficient}
Tomas Mikolov, Kai Chen, Greg Corrado, and Jeffrey Dean.
\newblock Efficient estimation of word representations in vector space.
\newblock \emph{arXiv preprint arXiv:1301.3781}, 2013.

\bibitem[Oord et~al.(2018)Oord, Li, and Vinyals]{oord2018representation}
Aaron van~den Oord, Yazhe Li, and Oriol Vinyals.
\newblock Representation learning with contrastive predictive coding.
\newblock \emph{arXiv preprint arXiv:1807.03748}, 2018.

\bibitem[Pedersen et~al.(2017)Pedersen, Stokholm, Thorsen, Mora-Jensen, and Bisgaard]{pedersen2017antibiotics}
Tine~Marie Pedersen, Jakob Stokholm, Jonathan Thorsen, Anna-Rosa~Cecilie Mora-Jensen, and Hans Bisgaard.
\newblock Antibiotics in pregnancy increase children's risk of otitis media and ventilation tubes.
\newblock \emph{The Journal of pediatrics}, 183:\penalty0 153--158, 2017.

\bibitem[Pichichero(2000)]{pichichero2000recurrent}
Michael~E Pichichero.
\newblock Recurrent and persistent otitis media.
\newblock \emph{The Pediatric infectious disease journal}, 19\penalty0 (9):\penalty0 911--916, 2000.

\bibitem[Qiu et~al.(2023)Qiu, Hu, Yuan, Zhou, Zhang, and Yang]{qiu2023not}
Zi-Hao Qiu, Quanqi Hu, Zhuoning Yuan, Denny Zhou, Lijun Zhang, and Tianbao Yang.
\newblock Not all semantics are created equal: Contrastive self-supervised learning with automatic temperature individualization.
\newblock \emph{arXiv preprint arXiv:2305.11965}, 2023.

\bibitem[Radford et~al.(2021)Radford, Kim, Hallacy, Ramesh, Goh, Agarwal, Sastry, Askell, Mishkin, Clark, et~al.]{radford2021learning}
Alec Radford, Jong~Wook Kim, Chris Hallacy, Aditya Ramesh, Gabriel Goh, Sandhini Agarwal, Girish Sastry, Amanda Askell, Pamela Mishkin, Jack Clark, et~al.
\newblock Learning transferable visual models from natural language supervision.
\newblock In \emph{International conference on machine learning}, pages 8748--8763. PmLR, 2021.

\bibitem[Stekhoven and B{\"u}hlmann(2012)]{stekhoven2012missforest}
Daniel~J Stekhoven and Peter B{\"u}hlmann.
\newblock Missforest—non-parametric missing value imputation for mixed-type data.
\newblock \emph{Bioinformatics}, 28\penalty0 (1):\penalty0 112--118, 2012.

\bibitem[Sundararajan et~al.(2017)Sundararajan, Taly, and Yan]{sundararajan2017axiomatic}
Mukund Sundararajan, Ankur Taly, and Qiqi Yan.
\newblock Axiomatic attribution for deep networks.
\newblock In \emph{International conference on machine learning}, pages 3319--3328. PMLR, 2017.

\bibitem[Tian et~al.(2020)Tian, Krishnan, and Isola]{tian2020contrastive}
Yonglong Tian, Dilip Krishnan, and Phillip Isola.
\newblock Contrastive multiview coding.
\newblock In \emph{Computer Vision--ECCV 2020: 16th European Conference, Glasgow, UK, August 23--28, 2020, Proceedings, Part XI 16}, pages 776--794. Springer, 2020.

\bibitem[Uno et~al.(2007)Uno, Cai, Tian, and Wei]{uno2007evaluating}
Hajime Uno, Tianxi Cai, Lu~Tian, and Lee-Jen Wei.
\newblock Evaluating prediction rules for t-year survivors with censored regression models.
\newblock \emph{Journal of the American Statistical Association}, 102\penalty0 (478):\penalty0 527--537, 2007.

\bibitem[Van~Buuren and Groothuis-Oudshoorn(2011)]{van2011mice}
Stef Van~Buuren and Karin Groothuis-Oudshoorn.
\newblock mice: Multivariate imputation by chained equations in r.
\newblock \emph{Journal of statistical software}, 45:\penalty0 1--67, 2011.

\bibitem[Van’T~Hof et~al.(2021)Van’T~Hof, Tisseur, van Berckelear-Onnes, Van~Nieuwenhuyzen, Daniels, Deen, Hoek, and Ester]{van2021age}
Maarten Van’T~Hof, Chanel Tisseur, Ina van Berckelear-Onnes, Annemyn Van~Nieuwenhuyzen, Amy~M Daniels, Mathijs Deen, Hans~W Hoek, and Wietske~A Ester.
\newblock Age at autism spectrum disorder diagnosis: A systematic review and meta-analysis from 2012 to 2019.
\newblock \emph{Autism}, 25\penalty0 (4):\penalty0 862--873, 2021.

\bibitem[Vaswani et~al.(2017)Vaswani, Shazeer, Parmar, Uszkoreit, Jones, Gomez, Kaiser, and Polosukhin]{vaswani2017attention}
Ashish Vaswani, Noam Shazeer, Niki Parmar, Jakob Uszkoreit, Llion Jones, Aidan~N Gomez, {\L}ukasz Kaiser, and Illia Polosukhin.
\newblock Attention is all you need.
\newblock \emph{Advances in neural information processing systems}, 30, 2017.

\bibitem[Walsh et~al.(2019)Walsh, McCormack, Webster, Pinto, Lee, Feng, Krakovsky, O’Grady, Tycko, Champagne, et~al.]{walsh2019maternal}
Kate Walsh, Clare~A McCormack, Rachel Webster, Anita Pinto, Seonjoo Lee, Tianshu Feng, H~Sloan Krakovsky, Sinclaire~M O’Grady, Benjamin Tycko, Frances~A Champagne, et~al.
\newblock Maternal prenatal stress phenotypes associate with fetal neurodevelopment and birth outcomes.
\newblock \emph{Proceedings of the National Academy of Sciences}, 116\penalty0 (48):\penalty0 23996--24005, 2019.

\bibitem[Xue et~al.(2020)Xue, Du, Mottram, Seneviratne, and Dai]{xue2020learning}
Yuan Xue, Nan Du, Anne Mottram, Martin Seneviratne, and Andrew~M Dai.
\newblock Learning to select best forecast tasks for clinical outcome prediction.
\newblock \emph{Advances in Neural Information Processing Systems}, 33:\penalty0 15031--15041, 2020.

\bibitem[Yao et~al.(2024)Yao, Yin, Cheung, Liu, and Qin]{yao2024drfuse}
Wenfang Yao, Kejing Yin, William~K Cheung, Jia Liu, and Jing Qin.
\newblock Drfuse: Learning disentangled representation for clinical multi-modal fusion with missing modality and modal inconsistency.
\newblock In \emph{Proceedings of the AAAI conference on artificial intelligence}, volume~38, pages 16416--16424, 2024.

\bibitem[Yuan et~al.(2021)Yuan, Lin, Kuen, Zhang, Wang, Maire, Kale, and Faieta]{yuan2021multimodal}
Xin Yuan, Zhe Lin, Jason Kuen, Jianming Zhang, Yilin Wang, Michael Maire, Ajinkya Kale, and Baldo Faieta.
\newblock Multimodal contrastive training for visual representation learning.
\newblock In \emph{Proceedings of the IEEE/CVF conference on computer vision and pattern recognition}, pages 6995--7004, 2021.

\bibitem[Yue et~al.(2022)Yue, Wang, Duan, Yang, Huang, Tong, and Xu]{yue2022ts2vec}
Zhihan Yue, Yujing Wang, Juanyong Duan, Tianmeng Yang, Congrui Huang, Yunhai Tong, and Bixiong Xu.
\newblock Ts2vec: Towards universal representation of time series.
\newblock In \emph{Proceedings of the AAAI conference on artificial intelligence}, volume~36, pages 8980--8987, 2022.

\end{thebibliography}

    \newpage
    \appendix
    \setcounter{equation}{0}
    \renewcommand{\theequation}{\thesection.\arabic{equation}}
    \setcounter{figure}{0}
    \renewcommand{\thefigure}{\thesection.\arabic{figure}}
    \setcounter{table}{0}
    \renewcommand{\thetable}{\thesection.\arabic{table}}

    \section{DTNN}
    \label{appx:dtnn} Here we follow the notation and setup proposed by \citet{hickey2024adaptive}. Let the observed survival dataset $\mathcal{D}= \{\vx_{i},y_{i}, s_{i}\}_{i=1}^{N}$, where $\vx_{i}$ is the input variable, and $y_{i}$ is a right-censoring time ($s_{i}= 0$) or an event time ($s_{i}=1$). Assuming non-informative right-censoring (e.g. no competing risk exists), DTNN categorizes time into different different time intervals including one additional class indicating censoring and calculates the risk for each time interval as multiclass classification problem \citep{lee2018deephit, hickey2024adaptive}. Following the assumption, we can optimize a DTNN model using the log-likelihood:

    \begin{equation}
        \cL_{\text{prediction}}= \sum_{i=1}^{N}s_{i}\log(\underbrace{p_{\theta}(y_i|x_i)}
        _{\text{assigned bin}}) + (1-s_{i})\log(\underbrace{S_{\theta}(y_i|x_i)}
        _{\text{sum over future bins}})
    \end{equation}

    \citet{hickey2024adaptive} provides a detailed procedure to calculate the
    loss function in the scenario where time intervals are fixed. Since the time-to-event prediction is converted to multi-class classification, the DTNN prediction head is simply a MLP with a softmax activation. When modeling survival data across $n$ time bins, the problem can be framed as a classification task with $n+1$ classes. The additional class corresponds to individuals who will not experience the event, commonly referred to as the ``cured'' instances in a mixture cure model \citep{boag1949maximum, farewell1982use}. Its probability is included in the survival $S_{\theta}(y_{i}|x_{i})$.

    \section{Forecasting Approach}
    \label{appx:forecast}

    Since $m_{\text{developmental}}$ is the most predictive modality and will dominate the prediction, an intuitive approach is to use past modality representations to forecast $m_{\text{developmental}}$'s representation. Following \citet{lyu2018improving} and \citet{xue2020learning}, we use an autoencoder to extract latent features from past data for forecasting $m_{\text{developmental}}$. The latent features are then used for final clinical prediction tasks.

    Since we only have one forecasting task, \citet{xue2020learning}'s meta-learning approach that automatically selecting forecasting tasks for representation learning reduces to the pretraining-and-finetuning paradigm in \citet{lyu2018improving}. In the pretraining step, a RNN encoder extracts latent features from representations of $m_{\text{prenatal}}$, $m_{\text{birth}}^{\text{mom}}$, and $m_{\text{birth}}^{\text{baby}}$ (for instance. mean pooling of CBOW token embeddings), and then a MLP decoder reconstructs the latent features into the representation of $m_{\text{developmental}}$. In the finetuning step, we use encoded latent features for the downstream task. Similar to the BFF approach, we use the same masking strategies to treat missing modality and enforce causal masking for prediction.

    \section{Multi-modal Self-Attention Scores}
    The softmax operation in the self-attention fusion applies the same masking strategies to treat missing modality and enforce causal masking for prediction. Eqn. \ref{eq:self-attn} is applied to either $\vz_{p}$ or $(\vz_{p}, \vz_{t})$ for each patient. Recall that $Z \in \mathbb{R}^{m\times d}$ is the multi-modal feature embedding matrix. The final representation is obtained as follows.
    \begin{equation}
        \vs = \text{masked-mean}\left( \text{masked-softmax}\left( \frac{QK^{\top}}{\sqrt{d_{k}}}
        \right)V \right)
        \label{eq:self-attn}
    \end{equation}
    where $Q = ZW^{Q}_{d \times d_k}$,
    $K = ZW^{K}_{d \times d_k}$, and $V = ZW^{V}_{d \times d_v}$.

    \label{sec:self-attn}
    \begin{figure}[!h]
        \centering
        \begin{minipage}{0.32\textwidth}
            \centering
            \includegraphics[width=\textwidth]{
                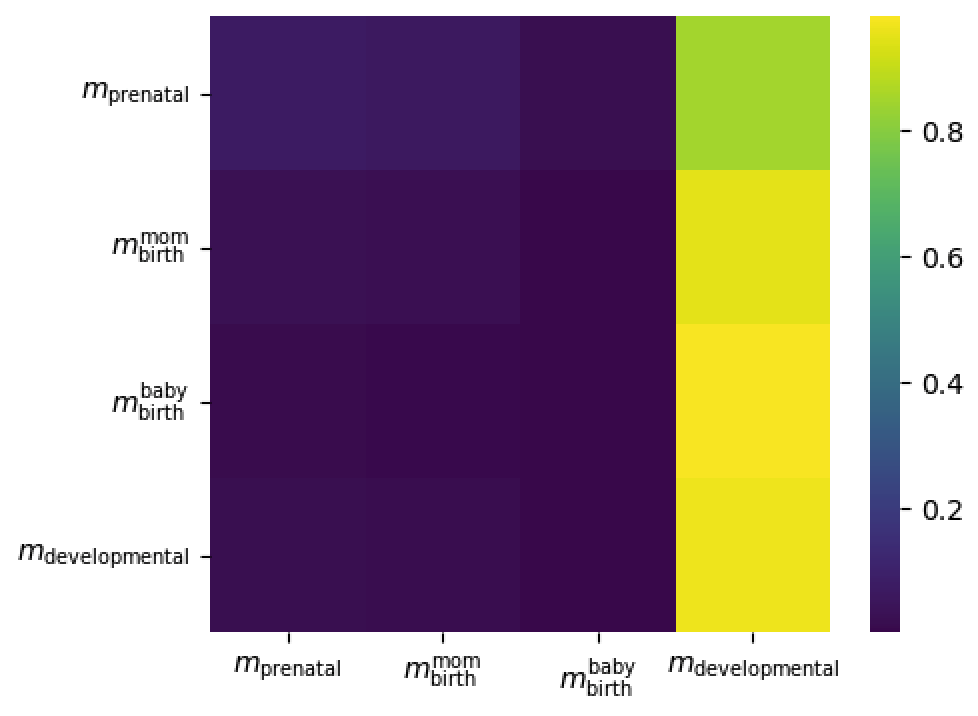
            }
        \end{minipage}
        \hfill
        \begin{minipage}{0.32\textwidth}
            \centering
            \includegraphics[width=\textwidth]{
                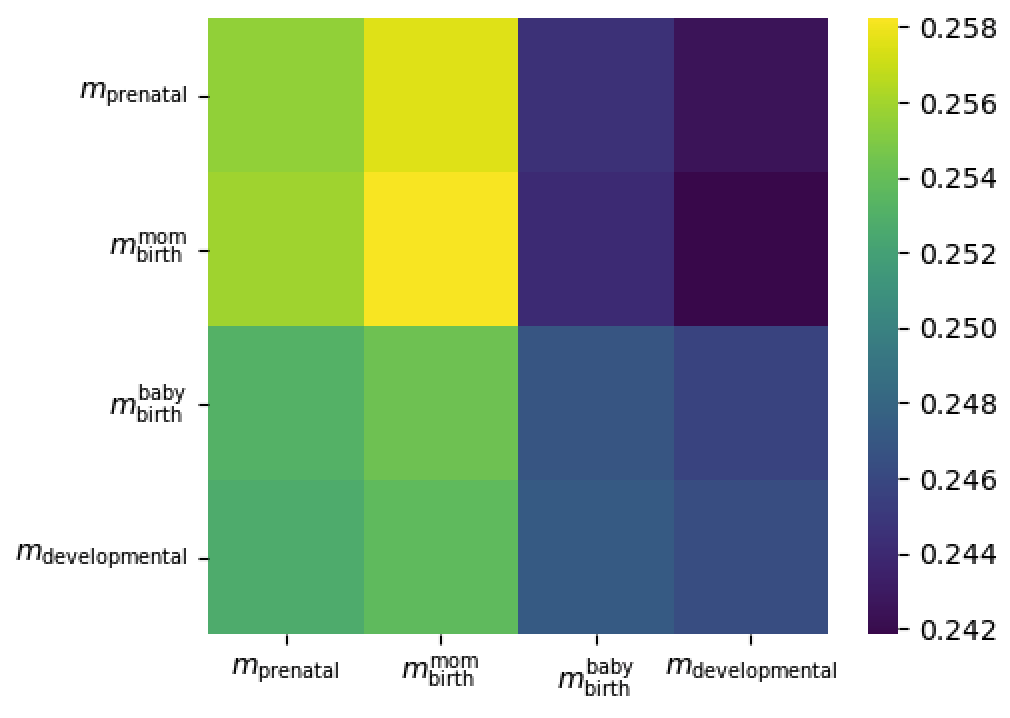
            }
        \end{minipage}
        \hfill
        \begin{minipage}{0.32\textwidth}
            \centering
            \includegraphics[width=\textwidth]{
                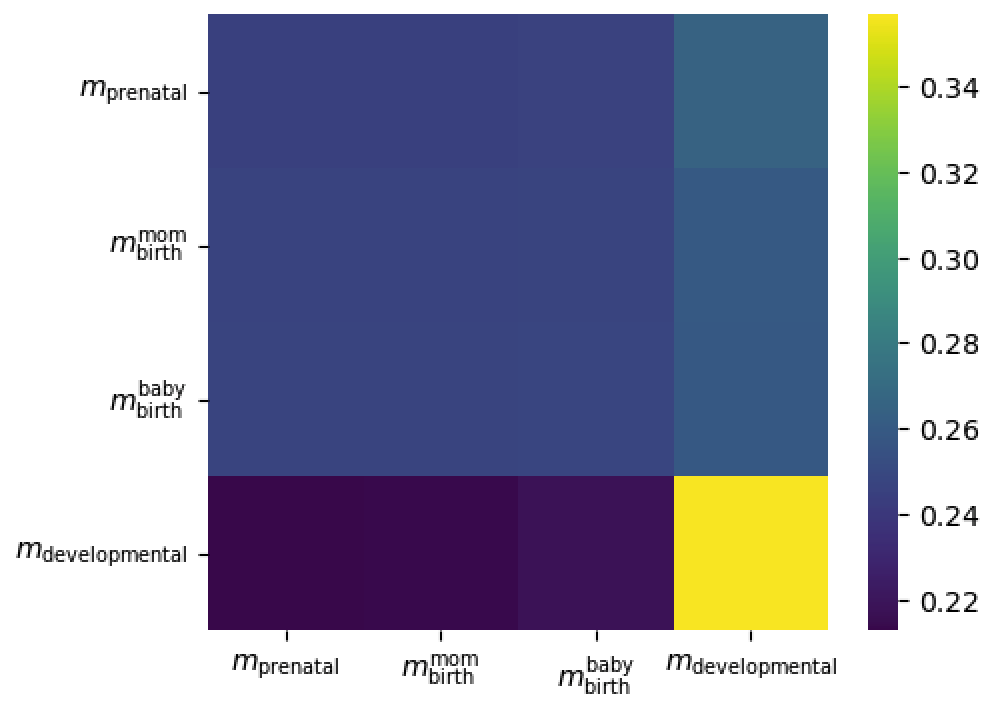
            }
        \end{minipage}

        \vspace{1em} 

        \begin{minipage}{0.32\textwidth}
            \centering
            \includegraphics[width=\textwidth]{
                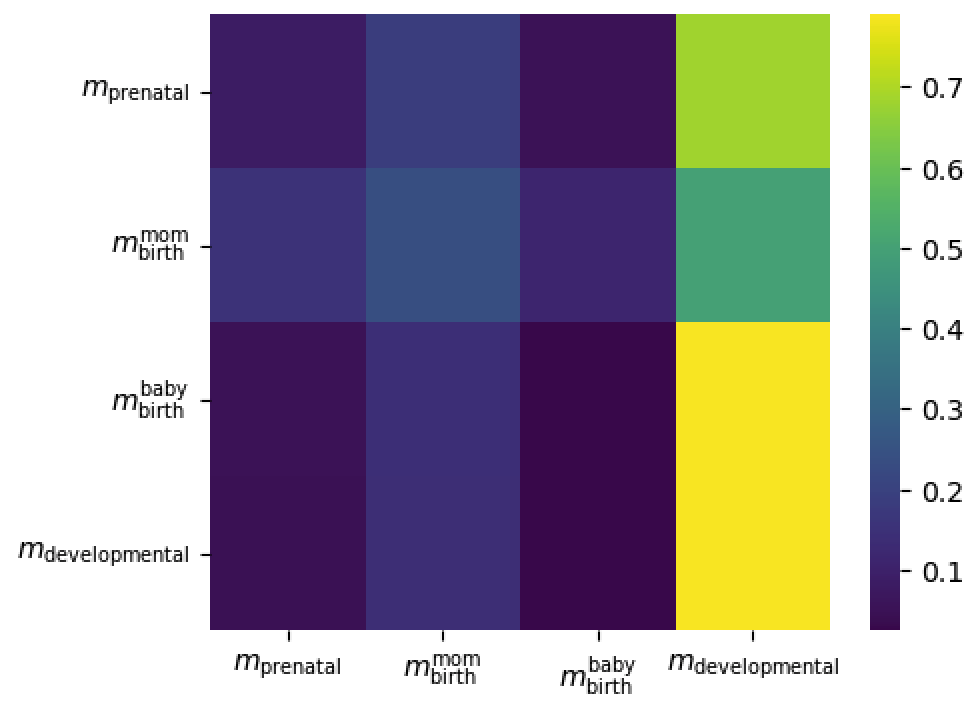
            }
            {\captionsetup{font=footnotesize}\caption*{(a) without CR}}
        \end{minipage}
        \hfill
        \begin{minipage}{0.32\textwidth}
            \centering
            \includegraphics[width=\textwidth]{
                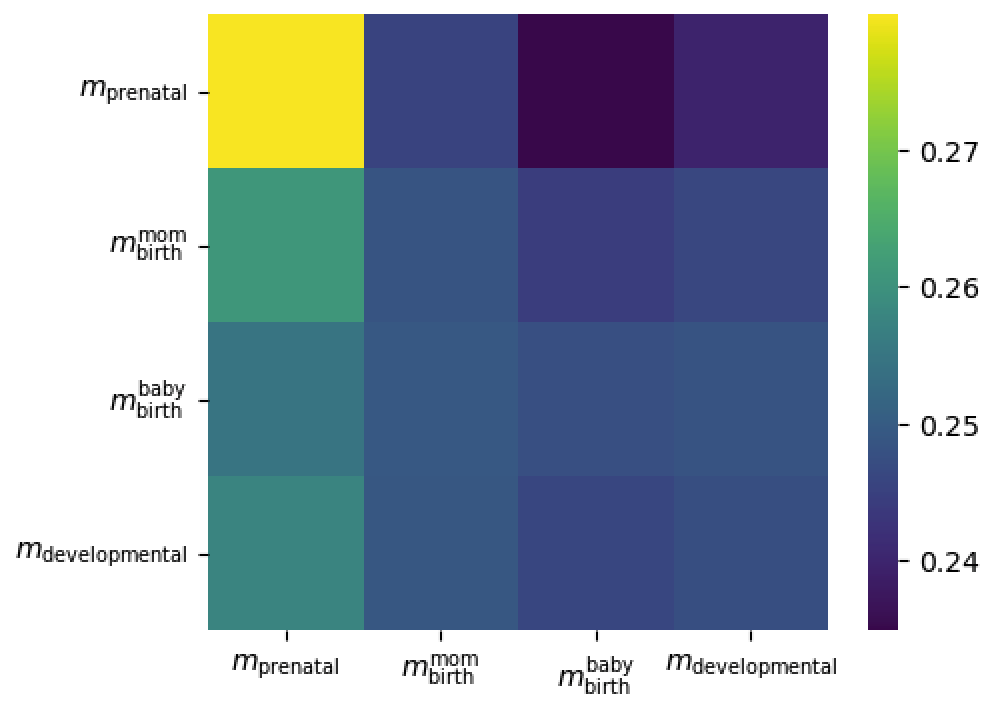
            }
            {\captionsetup{font=footnotesize}\caption*{(b) CR + $f(\vz_{p})$}}
        \end{minipage}
        \hfill
        \begin{minipage}{0.32\textwidth}
            \centering
            \includegraphics[width=\textwidth]{
                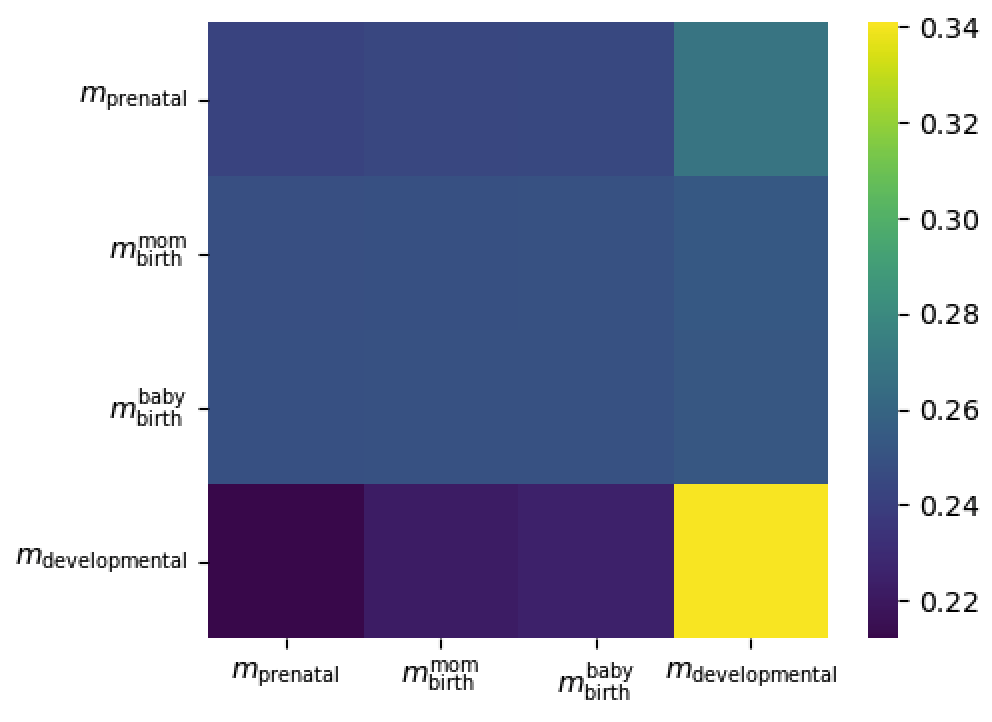
            }
            {\captionsetup{font=footnotesize}\caption*{(c) CR + $f(\vz_{p}, \vz_{t})$}}
        \end{minipage}

        \caption{Self-attention scores in the multi-modal fusion calculated using the modality representation (that is, either $\vz_p$ or ($\vz_p$, $\vz_t$)). The two samples presented here are identical to those used in Fig. \ref{fig:attn-self-gating}.}
        \label{fig:attn-self-attn}
    \end{figure}

    \section{IG-based modality importance}
    \label{appx:ig-percent} To calculate each modality's contribution, we compute the percentage of total absolute attributions that each modality receives during model training, using only samples where all modalities are fully observed. Specifically, we:
    \begin{enumerate}[label=(\arabic{*}), itemsep=0pt, parsep=0pt, topsep=0pt]
        \item Sum the absolute IG across the embedding dimension to get token-level attributions;

        \item Apply min-max normalization to scale these attributions to a uniform range;

        \item Sum the normalized scores in each modality to get modality-level attributions;

        \item Calculate the percentage contribution of each modality relative to the total attribution, averaged across all samples.
    \end{enumerate}
\end{document}